\documentclass[sigconf, nonacm, pdfa]{acmart}
\newcommand\vldbdoi{10.14778/3636218.3636234}
\newcommand\vldbpages{808 - 822}
\newcommand\vldbvolume{17}
\newcommand\vldbissue{4}
\newcommand\vldbyear{2023}
\newcommand\vldbauthors{\authors}
\newcommand\vldbtitle{\shorttitle} 
\newcommand\vldbavailabilityurl{https://github.com/HugoZHL/Hetu/tree/embedmem/tools/EmbeddingMemoryCompression}
\newcommand\vldbpagestyle{empty}

\usepackage[a-2b]{pdfx}
\usepackage{bm}
\usepackage[ruled]{algorithm2e}
\usepackage{multirow}
\usepackage{multicol}
\usepackage{array}
\usepackage{tikz}
\usepackage{url}
\usepackage{xurl}
\usepackage{ulem}
\newcommand*{\circled}[1]{\lower.7ex\hbox{\tikz\draw (0pt, 0pt)%
    circle (.5em) node {\makebox[1em][c]{\small #1}};}}

\usepackage{enumerate}

\usepackage{subfigure}
\usepackage{lipsum}

\usepackage{xcolor}
\usepackage{colortbl}
\usepackage{hhline}

\usepackage[utf8]{inputenc} 
\usepackage[T1]{fontenc}    
\usepackage{booktabs}       
\usepackage{amsfonts}       
\usepackage{nicefrac}       
\usepackage{microtype}      
\usepackage[utf8]{inputenc} 
\usepackage[T1]{fontenc}    
\usepackage{graphicx}
\usepackage{balance}  
\usepackage{soul}
\usepackage{bbm}
\usepackage[utf8]{inputenc}
\usepackage{amsmath}

\usepackage{amssymb}
\usepackage{amsthm}
\usepackage{booktabs}
\usepackage{amssymb}
\usepackage{makecell}

\usepackage{wrapfig}
\usepackage{hyperref}
\usepackage{enumitem}
\usepackage{float}
\usepackage{footnote}
\usepackage[textsize=small]{todonotes}

\usepackage{xspace}
\usepackage{amsmath}

\begin{document}
\title{Experimental Analysis of Large-scale Learnable Vector Storage Compression}

\author{Hailin Zhang}
\authornote{School of Computer Science \& Key Lab of High Confidence Software Technologies, Peking University}
\affiliation{%
  \institution{Peking University}
}
\email{z.hl@pku.edu.cn}

\author{Penghao Zhao}
\authornotemark[1]
\affiliation{%
  \institution{Peking University}
}
\email{penghao.zhao@stu.pku.edu.cn}

\author{Xupeng Miao}
\affiliation{%
  \institution{Carnegie Mellon University}
}
\email{xupeng@cmu.edu}

\author{Yingxia Shao}
\affiliation{%
  \institution{Beijing University of Posts and Telecommunications}
}
\email{shaoyx@bupt.edu.cn}

\author{Zirui Liu}
\authornotemark[1]
\affiliation{%
  \institution{Peking University}
}
\email{zirui.liu@pku.edu.cn}

\author{Tong Yang}
\authornotemark[1]
\affiliation{%
  \institution{Peking University}
}
\email{yangtongemail@gmail.com}


\author{Bin Cui}
\authornotemark[1]
\authornote{Institute of Computational Social Science, Peking University (Qingdao)}
\affiliation{%
  \institution{Peking University}
}
\email{bin.cui@pku.edu.cn}




\begin{abstract}


Learnable embedding vector is one of the most important applications in machine learning, and is widely used in various database-related domains. However, the high dimensionality of sparse data in recommendation tasks and the huge volume of corpus in retrieval-related tasks lead to a large memory consumption of the embedding table, which poses a great challenge to the training and deployment of models. Recent research has proposed various methods to compress the embeddings at the cost of a slight decrease in model quality or the introduction of other overheads. Nevertheless, the relative performance of these methods remains unclear. Existing experimental comparisons only cover a subset of these methods and focus on limited metrics. In this paper, we perform a comprehensive comparative analysis and experimental evaluation of embedding compression. We introduce a new taxonomy that categorizes these techniques based on their characteristics and methodologies, and further develop a modular benchmarking framework that integrates 14 representative methods. Under a uniform test environment, our benchmark fairly evaluates each approach, presents their strengths and weaknesses under different memory budgets, and recommends the best method based on the use case. In addition to providing useful guidelines, our study also uncovers the limitations of current methods and suggests potential directions for future research.

\end{abstract}

\maketitle

\pagestyle{\vldbpagestyle}
\begingroup\small\noindent\raggedright\textbf{PVLDB Reference Format:}\\
\vldbauthors. \vldbtitle. PVLDB, \vldbvolume(\vldbissue): \vldbpages, \vldbyear.\\
\href{https://doi.org/\vldbdoi}{doi:\vldbdoi}
\endgroup
\begingroup
\renewcommand\thefootnote{}\footnote{\noindent
This work is licensed under the Creative Commons BY-NC-ND 4.0 International License. Visit \url{https://creativecommons.org/licenses/by-nc-nd/4.0/} to view a copy of this license. For any use beyond those covered by this license, obtain permission by emailing \href{mailto:info@vldb.org}{info@vldb.org}. Copyright is held by the owner/author(s). Publication rights licensed to the VLDB Endowment. \\
\raggedright Proceedings of the VLDB Endowment, Vol. \vldbvolume, No. \vldbissue\ %
ISSN 2150-8097. \\
\href{https://doi.org/\vldbdoi}{doi:\vldbdoi} \\
}\addtocounter{footnote}{-1}\endgroup

\ifdefempty{\vldbavailabilityurl}{}{
\vspace{.3cm}
\begingroup\small\noindent\raggedright\textbf{PVLDB Artifact Availability:}\\
The source code, data, and/or other artifacts have been made available at \url{\vldbavailabilityurl}.
\endgroup
}

\section{Introduction}

In recent years, embedding techniques have
been widely used in database-related research areas, such as cardinality estimation~\cite{DBLP:journals/pvldb/LiuD0Z21,DBLP:journals/pvldb/KwonJS22}, query optimization~\cite{DBLP:journals/pvldb/ZhaoCSM22,DBLP:journals/pvldb/ChenGCT23}, language understanding~\cite{DBLP:journals/pvldb/KimSHL20}, entity resolution~\cite{DBLP:journals/pvldb/EbraheemTJOT18}, document retrieval~\cite{DBLP:journals/pvldb/HuangSLPKY20}, graph learning~\cite{DBLP:journals/pvldb/KochsiekG21,DBLP:journals/pvldb/YangSX0LB20} and advertising recommendation~\cite{DBLP:conf/sigmod/MiaoSZZNY022}. 
These applications, especially recommendation~\cite{DBLP:journals/dase/HanZYXCS23,DBLP:journals/dase/YuanZYHX23,DBLP:journals/jcst/HanWN23,DBLP:journals/chinaf/ShaoWCZW22} and retrieval~\cite{DBLP:journals/jcst/WuLMZM22,DBLP:conf/emnlp/KarpukhinOMLWEC20,DBLP:conf/naacl/QuDLLRZDWW21,DBLP:conf/emnlp/GaoC21}, often rely on large amount of embedding vectors to learn semantic representations and extract meaningful patterns and similarities.
However, the sheer volume of learnable vectors poses considerable storage challenges in practical deployment scenarios.
For example, Meta~\cite{DBLP:conf/isca/MudigereHHJT0LO22} proposed a deep learning recommendation model (DLRM) equipped with billions of embedding vectors that can take 96 terabytes memory to serve.

The management of these large amount of learnable vectors has become a critical concern for database communities (e.g., cloud-native vector database~\cite{DBLP:journals/pvldb/GuoLXYYLCXLLCQW22}).
One way to address the issue
is to involve multiple distributed instances, which may also bring significant communication overheads~\cite{DBLP:journals/pvldb/MiaoZSNYTC21,DBLP:conf/recsys/WangWLLYLLAGDSL22}.
Another way is to compress the embedding vectors without compromising the accuracy or the utility of models.
During the past few years, various compression methods have been proposed, including hashing, quantization, and so on. However, the performance and the effectiveness of these techniques remain largely unexplored. It is still an open question for data scientists to select from existing compression techniques when the storage of embeddings becomes unbearable.

\begin{figure}[htbp]
\includegraphics[width=0.84\linewidth]{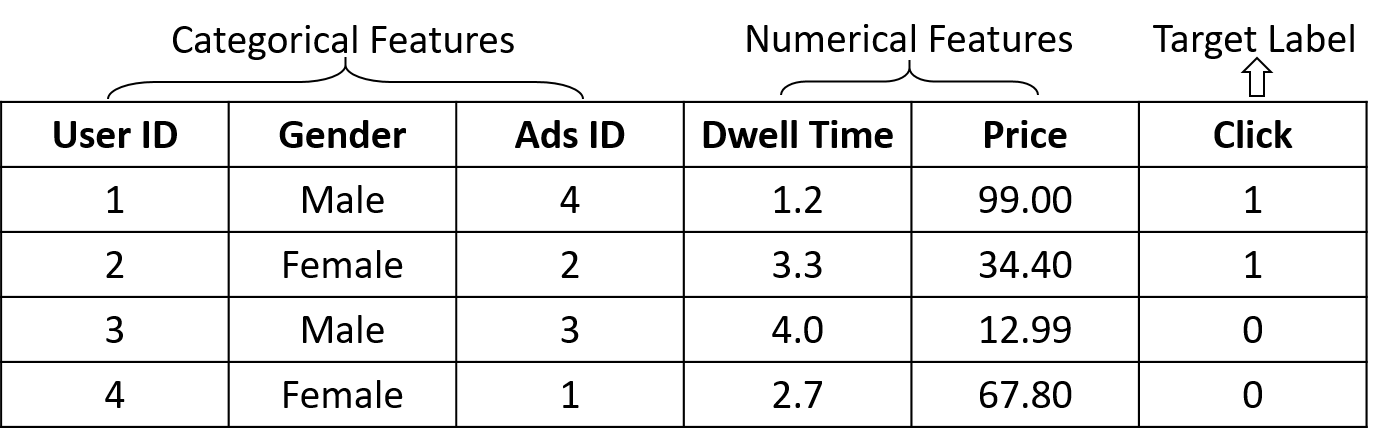}
\caption{An example of input data for DLRMs.}\label{fig:tabular}
\end{figure}

\begin{figure*}[htbp]
\includegraphics[width=\linewidth]{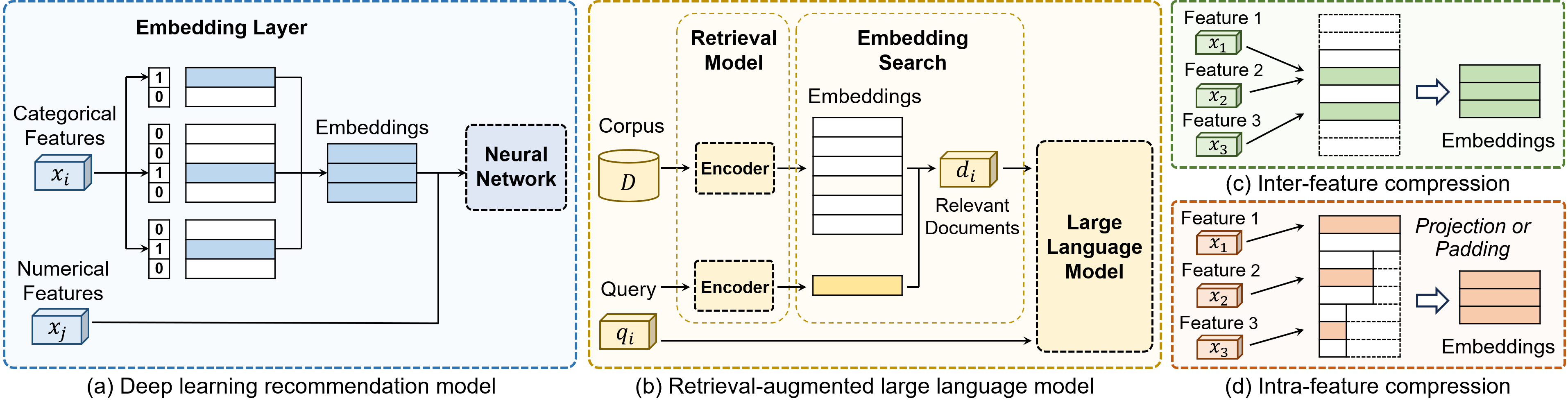}
\caption{(a) A typical DLRM. (b) A typical retrieval-augmented LLM. (c) An example of inter-feature compression, the original 8 features now share 4 embeddings. (d) An example of intra-feature compression, each embedding is compressed individually.}\label{fig:model}
\end{figure*}

In this paper, we study the above problem by revisiting the embedding compression methods under recommendation and retrieval scenarios since they have the most severe learnable vector storage pressure due to the high-dimensional sparse data~\cite{DBLP:conf/kdd/LianYZWHWSLLDLL22} and the huge volume of corpus. 
Figure~\ref{fig:tabular} illustrates an example of input data for DLRMs, which consists of multiple columns of categorical and numerical features, along with a column of target labels.
A typical DLRM vectorizes the categorical features into dense embeddings and feeds them along with numerical features into a downstream neural network to make predictions, as shown in Figure~\ref{fig:model}(a). 
The embedding layer maintains trainable embedding vectors for all categorical features.
It also provides embedding read and write primitives which are similar to the key-value storage~\cite{DBLP:conf/mlsys/ZhaoXJQDS020,DBLP:conf/cikm/ZhaoZXQJ019,DBLP:conf/sc/XieRLYXWLAXS20}.
Unfortunately, most existing key-value storage compression techniques~\cite{DBLP:journals/pvldb/RenZAG17,DBLP:conf/damon/ChenZXQLLZ21} are not suitable for DLRMs because of several special characteristics of embeddings,
such as skew distribution of embedding popularity~\cite{DBLP:conf/sigir/ZhangC15,DBLP:journals/corr/abs-2003-07336,DBLP:journals/pvldb/MiaoZSNYTC21,DBLP:conf/sigmod/MiaoSZZNY022,DBLP:conf/asplos/SethiAAKTW22,DBLP:conf/icde/ZhangCYYYZWDXSL22}, frequently accessing and updating of multiple embeddings especially during training. For example, if trained on the Criteo dataset, embeddings are accessed and updated more than 30 times per epoch, with the most popular embeddings being updated almost every iteration.
To address the memory issue, many embedding compression methods have been proposed and can be classified into two categories: inter-feature compression and intra-feature compression.

Considering that the storage bottleneck is mainly caused by the increasing number of unique features, \textbf{inter-feature compression forces features to share embeddings within a limited memory space}, as shown in Figure~\ref{fig:model}(c). Inter-feature compression is commonly used in industrial applications~\cite{DBLP:conf/mlsys/ZhaoXJQDS020}, and it requires an encoding function to maintain the mapping from features to embeddings. According to whether the encoding function is predetermined or updated during training, we further divide these methods into static encoding~\cite{DBLP:conf/kdd/ShiMNY20,DBLP:conf/recsys/ZhangLXKTGMDPIU20,DBLP:conf/mlsys/PansareKACSTV22} and dynamic encoding~\cite{DBLP:conf/www/KangCCYLHC20,adaptiveembedding,DBLP:journals/corr/abs-2302-01478}.

Inspired by features' different importance, \textbf{intra-feature compression assigns each feature an individually compressed embedding}.
Figure~\ref{fig:model}(d) shows an example, 
where each feature has its own embedding of distinct dimensions, and the final embeddings are obtained by projection or padding. 
According to the compression paradigm, intra-feature compression can be further divided into quantization, dimension reduction, and pruning. Quantization is a common compression method in deep learning models that uses data types with fewer bits~\cite{DBLP:conf/icml/GuptaAGN15,DBLP:conf/nips/LiD0SSG17}.
Dimension reduction~\cite{DBLP:conf/isit/GinartNMYZ21,DBLP:conf/www/ZhaoLLTGSWGL21,DBLP:conf/cikm/Lyu0ZG0TL22} and pruning~\cite{DBLP:conf/cikm/Lyu0ZG0TL22,kong2022autosrh} provide features with embeddings of different dimensions and sparsities, respectively.

Despite the existence of numerous proposed embedding compression methods, a thorough evaluation and analysis remains lacking. 
To the best of our knowledge, no previous work provides a comprehensive overview of this field. 
The experiments of existing approaches are often limited to specific cases with restricted metrics and settings. Consequently, the advantages, disadvantages, and applicability
of these compression methods have yet to be explored. While benchmarks for DLRMs have been established~\cite{mlperf,DBLP:conf/cikm/ZhuLYZH21}, they primarily focus on model design and do not consider embedding compression. 
The absence of a comprehensive evaluation framework for various compression methods makes it difficult to reproduce and compare existing techniques, which significantly undermines the practical value of research in this domain.

In addition to DLRMs, retrieval models also have large embedding tables for similarity-based embedding search. Although existing work focuses more on designing efficient embedding search algorithms~\cite{DBLP:conf/nips/SubramanyaDSKK19,DBLP:conf/nips/0015ZL20,DBLP:conf/nips/ChenZWLLLYW21}, the emergence of retrieval-augmented large language models (LLMs)~\cite{DBLP:conf/nips/LewisPPPKGKLYR020,DBLP:conf/icml/BorgeaudMHCRM0L22,guu2020retrieval} brings new challenges to embedding vector storage. A typical retrieval-augmented LLM is shown in Figure~\ref{fig:model}(b). Since LLMs already consume a lot of memory~\cite{DBLP:journals/jmlr/RaffelSRLNMZLL20,DBLP:conf/nips/BrownMRSKDNSSAA20}, embedding tables cannot be stored in GPUs or other accelerators, resulting in high search latency. It is currently unclear whether existing learnable vector compression methods are suitable for embeddings generated from retrieval models.

Motivated by the aforementioned issues in this research field, we aim to provide an in-depth analysis and a comprehensive experimental evaluation of embedding compression methods. 
In this paper, we carry out experiments using a unified evaluation framework to uncover the strengths and weaknesses of each method in various scenarios. 
We summarize our contributions as follows:
\begin{itemize}[leftmargin=*,parsep=0pt,itemsep=0pt,topsep=2pt,partopsep=2pt]
    \item We propose a new taxonomy of embedding compression methods according to their unique properties. On this basis, we provide a new perspective to understand and analyze their characteristics.
    \item We construct a unified modular evaluation framework for experiments. We build a general pipeline that can implement a wide variety of compression methods without much effort.
    \item We comprehensively evaluate representative embedding compression methods using rich metrics in DLRM scenarios. We further discuss the strengths and weaknesses of these methods.
    \item We apply the embedding compression methods to a retrieval-augmented LLM and analyze their performance.
    \item We discuss the guidelines, challenges, and promising research directions of embedding compression methods.
\end{itemize}


\section{Preliminaries}\label{sec:prelim}


\subsection{DLRM}\label{sec:prelim:dlrm}
The general architecture of DLRM is depicted in Figure \ref{fig:model}(a). A DLRM consists of two parts: an embedding layer mapping each categorical feature into a dense embedding vector, and a neural network containing interaction layers and fully connected layers. Numerical features are fed along with embeddings into the neural network. Many works have been done to improve the performance of the neural network part, such as WDL~\cite{DBLP:conf/recsys/Cheng0HSCAACCIA16}, DCN~\cite{DBLP:conf/kdd/WangFFW17}, and DIN~\cite{DBLP:conf/kdd/ZhouZSFZMYJLG18}.

In DLRMs, the categorical feature $x$ can be interpreted as a one-hot vector by encoding function $\mathcal{I}(x)$ to obtain the corresponding row vector $e$ from the embedding table $E\in\mathbb{R}^{n\times d}$ by $e=\mathcal{I}(x)^T E$; or $e=\mathcal{E}(\mathcal{I}(x))$ where $\mathcal{E}$ denotes the embedding layer function. Using $k$ to denote the number of categorical feature fields, and $x_{num}$ to denote numerical features, the downstream neural network is a function $f$ with parameters $\theta$ that inputs embeddings and outputs predictions $\hat{y} = f(e_{i_1},e_{i_2},...,e_{i_k},x_{num};\theta)$ for the loss function $\mathcal{L}$.
After the forward pass, optimizer such as Adam~\cite{DBLP:journals/corr/KingmaB14} is applied to update the embeddings and other model parameters. In summary, the optimization of DLRM can be formalized as:
\begin{equation}
\min\limits_{E,\theta} \mathbb{E}_{(X,y)\sim \mathcal{D}}\mathcal{L}(y,f(\mathcal{E}(\mathcal{I}(x)),x_{num};\theta)).
\end{equation}
The notations are detailed in Table \ref{tab:notations}.

\subsection{Retrieval-augmented LLM}\label{sec:prelim:rag}
The general structure of retrieval-augmented LLM is depicted in Figure~\ref{fig:model}(b). 
A typical model~\cite{DBLP:conf/nips/LewisPPPKGKLYR020,DBLP:conf/icml/BorgeaudMHCRM0L22,guu2020retrieval} consists of three parts: a retrieval model~\cite{DBLP:conf/emnlp/KarpukhinOMLWEC20,DBLP:conf/naacl/QuDLLRZDWW21,DBLP:conf/iclr/XiongXLTLBAO21}, an embedding search algorithm~\cite{DBLP:conf/nips/SubramanyaDSKK19,DBLP:conf/nips/0015ZL20,DBLP:conf/nips/ChenZWLLLYW21}, and an LLM~\cite{DBLP:conf/acl/LewisLGGMLSZ20,DBLP:journals/jmlr/RaffelSRLNMZLL20,DBLP:conf/nips/BrownMRSKDNSSAA20}. 
The retrieval model has two encoders $f_q$, $f_d$ that encode queries $q$ and all documents $D$ into embeddings separately. The embedding search algorithm $S$ takes query embedding $f_q(q)$ as input, and search similar documents within the embedding table $\mathcal{E}=f_d(D)$. After obtaining relevant documents $S(f_q(q_i),\mathcal{E})$, both the query and the documents serve as input to the LLM $f_{llm}$.
The size of the corpus used in industrial applications is at least one million level~\cite{DBLP:conf/nips/NguyenRSGTMD16,DBLP:journals/corr/abs-2309-13335}, resulting in a large amount of memory required for embedding table storage. For simplicity, we currently focus on the inference performance of retrieval-augmented LLMs.

\begin{table}[htb]
\centering
\small
\caption{Commonly used notations.}
\begin{tabular}{ c  l }

 \toprule[1pt]
 \textbf{Notation} & \textbf{Explanation} \\
 \midrule[0.8pt]
 $\mathcal{D}$ & Set of data samples  \\ 
 $E^{(*)}$ & Parameters of (compressed) embedding layer \\
 $\mathcal{E}^{(*)}$ & (Compressed) embedding layer  \\
 $k$ & Number of categorical fields  \\
 $X$ & Features of a sample \\
 $x, x_{num}$ & Categorical feature, numerical feature  \\
 $q$ & Input query for retrieval-augmented LLM \\
 $\mathcal{I}^{(*)}$ & (Compressed) encoding function \\
 $V$ & Set of features  \\
 $n$ & Number of features \\
 $m$ & Number of rows in inter-feature compression \\
 $d(')$ & (Reduced) embedding dimension \\
 $e$ & Embedding vector \\
 $r$ & Density in pruning methods \\
 $f,f_q,f_{llm}$ & Neural network of DLRM, query encoder, LLM \\
 $S$ & Embedding search function \\
 $Dec$ & Embedding decompress function \\
 $\theta$ & Model parameters except embeddings \\
 $y, \hat{y}$ & Ground truth label, prediction \\
 $\mathcal{L},\mathcal{L}_{llm}$ & Loss functions for DLRM and LLM \\
 $\mathcal{M},M_{budget}$ & Memory usage function, memory budget \\
 $CR$  & Compression ratio \\
 \bottomrule[1pt]
 \end{tabular}
\label{tab:notations}
\end{table}

\subsection{Problem Definition}\label{sec:prelim:def}
In this section, we discuss the problem of embedding compression in detail. 
In DLRMs, we use $\mathcal{E}^*$ to denote the compressed embedding layer with trainable parameters $E^*$. The encoding function $\mathcal{I}^*$ for the compressed embedding layer can be one-hot or multi-hot, depending on actual needs. The problem of learning the parameters of a DLRM with a compressed embedding layer can be modified to:
\begin{align}
\begin{split}
	\min\limits_{\mathcal{E}^*,\mathcal{I}^*,E^*,\theta} &\mathbb{E}_{(X,y)\sim \mathcal{D}}\mathcal{L}(y,f(\mathcal{E}^*(\mathcal{I}^*(x)),x_{num};\theta)), \\ 
	\mathrm{s.t.} &\mathcal{M}(\mathcal{E}^*,f) \le M_{budget}.
\end{split}
\end{align}
Besides the model parameters $E^*$ and $\theta$, the embedding layer $\mathcal{E}^*$ and the encoding function $\mathcal{I}^*$ are also variables that determine the loss. The optimization process can be decomposed into two parts: the first part determines $\mathcal{E}^*$ and $\mathcal{I}^*$ through the compression method, and the second part trains the model parameters $E^*$ and $\theta$. In this paper, our goal is to provide advice on choosing a proper compression method in the first part.

In retrieval-augmented LLMs, we use $Dec$ to denote the decompress function, and abuse some common notations such as $\mathcal{E}^*$ for compressed embeddings and $y$ for labels. Assuming our target is to minimize the objective function $\mathcal{L}_{llm}$, then the problem of choosing the most proper compression method can be formed as:
\begin{align}
\begin{split}
	\min\limits_{\mathcal{E}^*} &~\mathbb{E}_{q\sim \mathcal{D}}\mathcal{L}_{llm}(y,f_{llm}(q, S(f_q(q),Dec(\mathcal{E}^*)))), \\ 
	\mathrm{s.t.} &~\mathcal{M}(\mathcal{E}^*,f_{q},f_{llm},S) \le M_{budget}.
\end{split}
\end{align}
Since we are targeting the inference stage, the only variable is the compression algorithm. In practical applications, search algorithms are usually performed in batches, so the decompression can also be a batch operation to avoid storing the complete embedding table.

The compressed embeddings $\mathcal{E}^*$ should meet the memory constraint. The memory function $\mathcal{M}$ outputs the memory usage of the whole model during inference. In real scenarios, especially on-device situations, the memory budget is often smaller than the memory usage of models with the full embedding table. Since the memory usage of other parts is fixed, the memory constraint can be simplified as $\mathcal{M}(\mathcal{E}^*) \le M_{budget}$.
In addition to inference memory constraints, more metric constraints can be applied, such as low latency requirements in online service scenarios, training time or training memory constraints in time- or memory-limited scenarios.

Some methods cannot compress embedding layers within a given memory budget; they can only compress to a specific target memory. For instance, quantization methods directly adopt INT8 or INT16 to replace the original FLOAT32 data type, reducing the memory usage to 25\% or 50\%. To measure the compression ability, we define $CR$ (compression ratio) as the ratio of the original memory to the compressed memory as
$CR = \frac{\mathcal{M}(\mathcal{E})}{\mathcal{M}(\mathcal{E}^*)}$.

\subsection{Scope}\label{sec:prelim:scope}

In this section, we discuss the scope of this paper. 
We focus on embedding compression for DLRMs and retrieval-augmented LLMs, with practical applications of at least millions of embeddings.
We do not focus on the embeddings in NLP models. Although works have been done to compress the memory usage of embedding tables in NLP~\cite{DBLP:conf/nips/SvenstrupHW17,DBLP:conf/nips/ChenSLCH18,DBLP:conf/iclr/ShuN18,DBLP:conf/icml/ChenMS18,DBLP:conf/icml/ChenLS20}, the number of unique vocabularies in mainstream LLMs such as Bert~\cite{DBLP:conf/naacl/DevlinCLT19} and GPT-3~\cite{DBLP:conf/nips/BrownMRSKDNSSAA20} is no more than 0.1 million, which is not as memory-intensive as DLRMs. 

There are several research directions that can be easily confused with our study: feature selection, embedding search. Although feature selection~\cite{DBLP:conf/www/WangZ0W22,DBLP:conf/kdd/LinZW0W22,DBLP:journals/corr/abs-2301-10909} does reduce memory usage by directly pruning useless features, it can be seen as the upstream process of embedding compression. Embedding search~\cite{DBLP:journals/is/MalkovPLK14,DBLP:journals/pami/MalkovY20,DBLP:journals/tbd/JohnsonDJ21} is a subsequent stage of embedding compression in retrieval tasks, and its related research is orthogonal to memory compression.

\section{Overview of Embedding Compression}\label{sec:methods}

In this section, we present an overview and a new taxonomy of embedding compression methods. We first divide all methods into inter-feature and intra-feature compression based on whether the features share parameters or have individually compressed embeddings. We further divide the methods according to their properties and techniques. The detailed information of inter-feature and intra-feature compression methods is listed in Table~\ref{tab:interfeature} and \ref{tab:intrafeature}, respectively.

\begin{table*}[ht]
\small
\caption{Summary of inter-feature compression. Space Complexity reflects the memory of encoding function and embedding layer; Time Complexity reflects the time of embedding lookup process; Freq / Impo-aware indicates whether the method is frequency- or importance-aware; Com Cap shows the compression capability within a certain range of memory budgets.}\label{tab:interfeature}
\begin{tabular}{|p{1.6cm}<{\centering}|p{2.1cm}<{\centering}|p{2.2cm}<{\centering}|p{2.5cm}<{\centering}|p{2.5cm}<{\centering}|p{2.5cm}<{\centering}|p{1.6cm}<{\centering}|}
\hline
\textbf{Subcategory} & \textbf{Method} & \textbf{Techniques} & \textbf{Space Complexity} & \textbf{Time Complexity} & \textbf{Freq / Impo-aware} & \textbf{Com Cap} \\
\hline
\multirow{8}{*}{\makecell[c]{Static\\Encoding}} & CompoEmb~\cite{DBLP:conf/kdd/ShiMNY20} & Hash & $O(md+nd/m)$ & $O(d)$ & /  & Yes  \\
\cline{2-7}
& DoubleHash~\cite{DBLP:conf/recsys/ZhangLXKTGMDPIU20} & Hash & $O(md)$ & $O(d)$ & Frequency-aware &  Yes  \\
\cline{2-7}
& BinaryCode~\cite{DBLP:conf/cikm/YanWLLLXZ21} & Hash & $O(\sqrt{n}\cdot d)$ & $O(d)$ & /  & No  \\
\cline{2-7}
& MemCom~\cite{DBLP:conf/mlsys/PansareKACSTV22} & Hash & $O(md+n)$ & $O(d)$ & / &  Yes  \\
\cline{2-7}
& DHE~\cite{DBLP:conf/kdd/KangCYYCHC21} & Hash, MLP & $O(d\cdot d_i+ d_i^2)$ & $O(d\cdot d_i+ d_i^2)$ & / & Yes \\
\cline{2-7}
& TT-Rec~\cite{DBLP:conf/mlsys/YinAWL21} & TensorTrain (Hash) & $O(R^2\sum(m_i+d_i))$ & $O(R^2d)$ & Frequency-aware &  Yes  \\
\cline{2-7}
& ROBE~\cite{DBLP:conf/mlsys/DesaiCS22} & Hash, 1-D Array & $O(md)$ & $O(d)$ & / &  Yes  \\
\cline{2-7}
& Dedup~\cite{DBLP:journals/pvldb/ZhouCDMYZZ22} & LSH & $O(n+md)$ & $O(d)$ & / & Yes \\
\hline
\multirow{4}{*}{\makecell[c]{Dynamic\\Encoding}} & MGQE~\cite{DBLP:conf/www/KangCCYLHC20} & VQ (PQ) & $O(nk'+md)$ & $O(k'+d)$ & Frequency-aware & Yes  \\
\cline{2-7}
& LightRec~\cite{DBLP:conf/www/LianWLLC020} & VQ (AQ) & $O(nk'+md)$ & $O(k'd)$ & /  & Yes \\
\cline{2-7}
& AdaptEmb~\cite{adaptiveembedding} & Hash, Frequency & $O(m+md)$ & $O(d)$ & Frequency-aware &  Yes  \\
\cline{2-7}
& CEL~\cite{DBLP:journals/corr/abs-2302-01478} & Clustering & $O(n+2B/b \cdot d)$ & $O(d)$ & Frequency-aware  & Yes  \\
\hline
\end {tabular}
\end {table*}

\subsection{Inter-feature Compression}
To address the memory bottleneck caused by the explosive growth of features, a direct approach is to keep only a small number of embeddings for features to share, as shown in Figure~\ref{fig:model}(c). 
Compression is generally performed within fields to ensure that features which share embeddings have similar semantics.
Inter-feature compression needs to maintain a new mapping from features to embeddings, instead of the original one-hot encoding.
Early methods utilize hash functions~\cite{DBLP:conf/icml/WeinbergerDLSA09,DBLP:conf/nips/SvenstrupHW17} to map features into multi-hot vectors, then lookup from hash embedding tables for sub-embeddings to construct final embeddings. Following this idea, the problem can be simplified as finding an encoding function $\mathcal{I}^*$, and a corresponding row-compressed embedding layer $\mathcal{E}^*$. Based on whether the encoding function is fixed during training, the methods can be further divided into static encoding and dynamic encoding.

\subsubsection{Static Encoding}
Static encoding uses fixed encoding functions during training. Mapping features into a smaller number of embeddings is essentially a hashing process. Thus, many hash functions have served as encoding functions in industry~\cite{DBLP:conf/mlsys/ZhaoXJQDS020}. While early works explored the form of encoding functions, recent works further explored the form of embedding layers. 
We use $m$ to denote the number of embeddings after compression.

\noindent\textbf{DoubleHash}~\cite{DBLP:conf/recsys/ZhangLXKTGMDPIU20} uses two hash functions, and sums the two sub-embeddings together. More hash functions lead to less collision rate since the bucket size is enlarged from $m$ to $m^2$.

\noindent\textbf{CompoEmb}~\cite{DBLP:conf/kdd/ShiMNY20} recursively divides the original feature index by the row sizes of hash embedding tables and gets the remainders as the new indices.
As long as the product of row sizes is greater than the number of features, no features will share the exact same embedding.
The sub-embeddings are aggregated by multiplication.
BinaryCode~\cite{DBLP:conf/cikm/YanWLLLXZ21} follows the idea, splitting the binary respresentation of the original index
in succession style or skip style, where the former is essentially the CompoEmb. Some following work~\cite{DBLP:conf/cikm/LiCZY21} also adopts CompoEmb to implement lightweight embedding layers.

\noindent\textbf{MEmCom}~\cite{DBLP:conf/mlsys/PansareKACSTV22} stores scale and bias weights for each feature.
Given a feature as input, a embedding is indexed by a hash function, then multiplied and added with scale and bias to get the final embedding. 

Methods above share some sub-embeddings among features, resulting in degraded model quality. 
They only form the encoding function, enabling simple and flexible memory compression. 
In contrast, the following methods design new embedding layers.

\noindent\textbf{DHE} (Deep Hash Embedding~\cite{DBLP:conf/kdd/KangCYYCHC21}) radically replaces embedding tables with multi-layer perceptrons (MLPs). It maps features to integers using many hash functions and then applies transformations to approximate a uniform or Gaussian distribution as input to MLPs.
Equipped with complex MLPs, DHE achieves good model quality, but requires much more time to train and infer.
In Table~\ref{tab:interfeature}, the symbol $d_i$ means the number of hidden units in each MLP layer.

\noindent\textbf{TT-Rec} (Tensor-Train Recommendation~\cite{DBLP:conf/emnlp/HrinchukKMOO20,DBLP:conf/mlsys/YinAWL21}) borrows the idea of tensor-train decomposition (abbreviated as TT).
In TT, a tensor $\mathcal{A}\in\mathbb{R}^{I_1\times I_2\times...I_t}$ can be decomposed $\mathcal{A}\approx\mathcal{G}_1\mathcal{G}_2...\mathcal{G}_t$, with each TT-core $\mathcal{G}_i\in\mathbb{R}^{R_{i-1}\times I_i\times R_i}, R_0=R_t=1$. 
TT-Rec factorizes the row size $|V|\le\Pi_{i=1}^t m_i$ and the column size $d\le\Pi_{i=1}^t d_i$ into intergers, and decomposes the embedding table by $E\approx\mathcal{G}_1\mathcal{G}_2...\mathcal{G}_t$, where $\mathcal{G}_i\in\mathbb{R}^{R_{i-1}\times m_i\times d_i\times R_i}$. To obtain the embedding, TT-Rec looks up the tensors and conducts matrix multiplication according to the decomposition. 
Its row decomposition is the same with CompoEmb, but
the matrix multiplication requires much more time than simple aggregation. 
TT-Rec is also adopted by following work~\cite{DBLP:conf/www/WangYCHWZH20}. 

\noindent\textbf{ROBE} (Random Offset Block Embedding~\cite{DBLP:conf/mlsys/DesaiCS22}) stores an 1-D array instead of 2-D matrix for embedding layer. It uses hash functions to generate indices, then concatenates the sub-embeddings retrieved at the indices. 
ROBE can reduce running time with simple design, but requires more epochs to converge due to the randomness.

The exploration of embedding layer design is both creative and effective. They either improve the model quality at the cost of more computation, or simplify the embedding structure.

\noindent\textbf{Dedup}~\cite{DBLP:journals/pvldb/ZhouCDMYZZ22} conducts similarity-based deduplication~\cite{DBLP:conf/sigmod/VartakTMZ18,DBLP:conf/mobisys/LeeN20} on embedding models. It adopts L2LSH~\cite{DBLP:conf/stoc/IndykM98}, a local-sensitive hashing algorithm on Euclidean (L2) distance, to efficiently deduplicate similar parameter blocks.
Dedup can only be applied when the parameters are fixed, so uncompressed embeddings still need to be trained. We classify Dedup as static encoding, because the encoding function is determined by an one-pass LSH process. 
Since it deduplicates embeddings by value, we directly deduplicate the entire embedding table to speed up compression and serving, regardless of feature fields.
Dedup hashes the embedding content while other hashing-based methods hash the input indices, so we distinguish them explicitly in our experimental analysis.

In summary, static encoding methods are simple, effective, and capable of compression at any memory budget. Their encoding functions, which remain constant during training, are often simple hash functions that take up no storage space. The focus of research work gradually shifts from hash functions to embedding layers. The former guarantees memory constraints while the latter further guarantees model quality. Another line of research is similarity-based deduplication, which performs post-training compression.

\subsubsection{Dynamic Encoding}
Dynamic encoding allows encoding functions to be updated during training, which is naturally suitable for online learning. 
They adopt trainable indices or build data structures to store and adjust the mapping. 
They tend to incorporate more information but only achieve mediocre compression ratios. 

\noindent\textbf{MGQE} (Multi-Granularity Quantized Embedding~\cite{DBLP:conf/www/KangCCYLHC20}) extends DPQ (Differentiable Product Quantization)~\cite{DBLP:conf/icml/ChenLS20} to fit recommendation data. DPQ is based on PQ (Product Quantization)~\cite{DBLP:journals/pami/JegouDS11}, a VQ (Vector Quantization) technique in embedding search. PQ splits embeddings into several parts, clusters the partial embeddings respectively, then reconstructs embeddings with the nearest centroids.
DPQ introduces supervised learning to train the centroids, minimizing the distances between the original and the reconstructed embeddings.
The uncompressed embeddings are kept during training to determine and update the nearest sub-embeddings. After training, the uncompressed embeddings are dropped, and the nearest sub-embeddings are adopted to reconstruct the final embeddings.
DPQ and other similar works~\cite{DBLP:conf/iclr/ShuN18,DBLP:conf/icml/ChenMS18} focus on NLP word embeddings, and MGQE extends DPQ for highly-skewed recommendation data, providing more centroid embeddings for features with higher frequency.
The memory usage can only be reduced during inference, and the compression ratio is relatively low due to the storage of centroids indices. 
Besides PQ, other VQ techniques such as AQ (Additive Quantization) are also adopted for embedding compression~\cite{DBLP:conf/www/LianWLLC020}. The centroids in AQ are summed to reconstruct embeddings. In Table~\ref{tab:interfeature}, $k'$ is the number of parts in VQ.

\noindent\textbf{AdaptEmb} (DeepRec Adaptive Embedding~\cite{adaptiveembedding}) allocates unique embeddings for high-frequency features and shared embeddings for others.
It dynamically converts the feature's embedding from shared to exclusive if the frequency becomes high enough.
It incurs extra memory to store high frequency features during inference.

\noindent\textbf{CEL} (Clustered Embedding Learning~\cite{DBLP:journals/corr/abs-2302-01478}) compresses the embeddings of two special fields (users and items) that are clustered with only one embedding per cluster. During training, items are dynamically reassigned to the more proper clusters based on their history interactions, and clusters are split if associated with too many interactions.
CEL has limited compression ratios with the storage of the cluster structure, and takes more training time due to cluster adjustment. 
In Table~\ref{tab:interfeature}, the total number of interactions is $B$ and the cluster will split iff it has more than $2b$ associated interactions.

Dynamic encoding requires extra data structures to store dynamic codes, so sometimes only supports limited compression ratios. They incorporate frequency information, but the dynamic encoding function brings some overheads during training.

\begin{table*}[ht]
\small
\caption{Summary of inter-feature compression. Space Complexity reflects the memory of encoding function and embedding layer; Time Complexity reflects the time of embedding lookup process; Freq / Impo-aware indicates whether the method is frequency- or importance-aware; Com Cap shows the compression capability within a certain range of memory budgets.}\label{tab:intrafeature}
\begin{tabular}{|p{1.6cm}<{\centering}|p{2.1cm}<{\centering}|p{2.2cm}<{\centering}|p{2.5cm}<{\centering}|p{2.5cm}<{\centering}|p{2.5cm}<{\centering}|p{1.6cm}<{\centering}|}
\hline
\textbf{Subcategory} & \textbf{Method} & \textbf{Techniques} & \textbf{Space Complexity} & \textbf{Time Complexity} & \textbf{Freq / Impo-aware} & \textbf{Com Cap} \\
\hline
\multirow{5}{*}{Quantization} & FP16~\cite{zhang2018training} & FP16 & $O(nd/2)$ & $O(d)$ & /  & No  \\
\cline{2-7}
& Post4Bits~\cite{DBLP:journals/corr/abs-1911-02079} & Greedy Search & $O(nd/8)$ & $O(d)$ & /  & No  \\
\cline{2-7}
& MixedPrec~\cite{DBLP:journals/corr/abs-2010-11305} & FP16, INT8 & $O(nd/4)$ & $O(d)$ & / & No  \\
\cline{2-7}
& Int8/16~\cite{DBLP:conf/sigmod/XuLZSHL021} & INT8, INT16 & $O(nd/4)$ & $O(d)$ & / & No \\
\cline{2-7}
& ALPT~\cite{DBLP:journals/corr/abs-2212-05735} & Learnable Scale & $O(nd/4 + n)$ & $O(d)$ & Importance-aware & No  \\
\hline
\multirow{8}{*}{\makecell[c]{Dimension\\Reduction}} & NIS~\cite{DBLP:conf/kdd/JoglekarLCXWAKL20} & Policy Gradient & $O(nd')$ & $O(d'·d)$ & Importance-aware & Yes \\
 \cline{2-7}
 & ESAPN~\cite{DBLP:conf/sigir/LiuZWLT20} & Policy Gradient & $O(nd')$ & $O(d'·d)$ & Both & No \\
 \cline{2-7}
 & MDE~\cite{DBLP:conf/isit/GinartNMYZ21} & Heuristic & $O(nd')$ & $O(d'·d)$ & Frequency-aware & Yes  \\
 \cline{2-7}
 & AMTL~\cite{DBLP:conf/cikm/YanWZLLXZ21} & MLP & $O(nd')$ & $O(d)$ & Both & No  \\
 \cline{2-7}
 & AutoEmb~\cite{DBLP:conf/icdm/ZhaoLFLTWCZLY21} & DARTS & $O(nd')$ & $O(d'·d)$ & Both  & No \\
 \cline{2-7}
 & AutoDim~\cite{DBLP:conf/www/ZhaoLLTGSWGL21} & DARTS & $O(nd')$ & $O(d'·d)$ & Importance-aware  & No  \\
 \cline{2-7}
 & SSEDS~\cite{DBLP:conf/sigir/QuYTZSY22} & One-shot NAS & $O(nd')$ & $O(d'·d)$ & Importance-aware  & Yes  \\
 \cline{2-7}
 & OptEmbed~\cite{DBLP:conf/cikm/Lyu0ZG0TL22} & One-shot NAS & $O(nd')$ & $O(d)$ & Importance-aware  & No  \\
\hline
\multirow{5}{*}{Pruning} & DeepLight~\cite{DBLP:conf/wsdm/0002PZKFL21} & Structural Prune & $O(rnd)$ & $O(rd)$ & Importance-aware & Yes  \\
\cline{2-7}
& PEP~\cite{DBLP:conf/iclr/LiuGCJL21} & Mask, Threshold & $O(rnd)$ & $O(rd)$ & Importance-aware & No  \\
\cline{2-7}
& HAM~\cite{DBLP:conf/cikm/XiaoXCC22} & STE, Hard Mask & $O(rnd)$ & $O(rd)$ & Importance-aware & Yes \\
\cline{2-7}
& AutoSrh~\cite{kong2022autosrh} & Mask, DARTS & $O(rnd)$ & $O(rd)$ & Both & Yes  \\
\hline
\end {tabular}
\end {table*}

\subsection{Intra-feature Compression}
Instead of sharing embeddings and modifying the encoding function, intra-feature compression compresses embeddings individually to form a new embedding layer $\mathcal{E}^*$. These methods can be further divided into quantization, dimension reduction, pruning.

\subsubsection{Quantization}
Quantization is a common compression technique in deep learning training~\cite{DBLP:journals/jmlr/HubaraCSEB17,DBLP:conf/iclr/MicikeviciusNAD18} and inference~\cite{DBLP:conf/nips/BannerNS19}. 
It is stable and simple to use, since it does not affect the original training paradigm; however, low-precision data types will lead to a slight loss of model quality and limited compression ratios.

\noindent\textbf{FP16}~\cite{zhang2018training} 
is only used for storage, while during training the retrieved embeddings are converted to FP32.
When rounding updated parameters back into FP16, there are two choices: nearest rounding and stochastic rounding. The former selects the nearest value in FP16, appearing to have a systematic bias in model update accumulation since a relatively small update will never take effect if always discarded. The latter first computes the rounded-up value and the rounded-down value, then draws a random number from a Bernoulli distribution with the distances to these values; this approach is not biased yet brings higher variance in optimizer. In practice, stochastic rounding is chosen for better model quality.

\noindent\textbf{INT8/16}~\cite{DBLP:journals/corr/abs-2010-11305,DBLP:conf/sigmod/XuLZSHL021} 
are integer data types of low-precision, treated as bins of values, where two manually designed parameters scale and  bias are further required to restore the original FP32 value.
FP data types have unequal intervals between values, while INT data types have equal intervals.
INT data types also adopt stochastic rounding for better model quality.

Directly using FP16 or INT8/16 is simple and has almost no overhead. In order to obtain better model quality, the following methods try to search or learn the proper scale for INT-type compression.

\noindent\textbf{Post4Bits}~\cite{DBLP:journals/corr/abs-1911-02079} performs a post-training greedy search on scale and bias of INT4 data type. The minimum and the maximum values are searched step by step to minimize the model loss. 

\noindent\textbf{ALPT}~\cite{DBLP:journals/corr/abs-2212-05735} makes the scale alternatively trained with the model parameters to improve the model quality. The idea of learnable scale comes from LSQ~\cite{DBLP:conf/iclr/EsserMBAM20}. 

In summary, quantization involves little overhead and achieves certain compression ratios.

\subsubsection{Dimension Reduction}
Generally, the larger the dimension, the more information the embedding can represent. As the recommendation data is highly skewed
~\cite{DBLP:conf/sigir/ZhangC15}, a natural idea is to assign different dimensions for features with different frequency or importance. To align the dimension of embeddings for subsequent neural networks, there are two ways: zero-padding and projection. The second scheme is inspired by SVD~\cite{DBLP:conf/nips/ChenSLCH18} and is adopted by most methods, since the learnable projection matrices can represent all the linear transformations including zero-padding. The symbol $d'$ in Table~\ref{tab:intrafeature} means the reduced dimension.

\noindent\textbf{MDE} (Mixed Dimension Embedding~\cite{DBLP:conf/isit/GinartNMYZ21}) 
represents feature frequency with the inverse of feature cardinality within field. 
The dimensions are proportional to $p^\alpha$, where $p$ is the frequency and $\alpha$ is a hyper-parameter. 

MDE is the only dimension reduction method that compresses memory during training, with no learnable structures involved. All of the following methods adopt learnable structures to determine dimensions, incurring much more training overhead.

\noindent\textbf{NIS} (Neural Input Search~\cite{DBLP:conf/kdd/JoglekarLCXWAKL20}) uses a policy network to determine the dimensions. It splits the embeddings into chunks. For each column chunk, it builds projection matrices and uses a controller to sample row chunks. In the reward $R=R_Q-\lambda\cdot C_M$, $R_Q$ is the model quality and $C_M$ is the memory cost at inference. 
The chunk-based search is also used in~\cite{DBLP:conf/kdd/ChenYZHWW21}.

\noindent\textbf{ESAPN} (Embedding Size Adjustment Policy Network~\cite{DBLP:conf/sigir/LiuZWLT20}) uses a series of projection matrices to convert dimensions larger and larger until the final dimension. Each feature field is assigned a policy network, which inputs the feature frequency and the current dimension and outputs whether enlarge the dimension. 
If the dimension is enlarged, the transformed vector is used as initialization. It takes the improvement of the current state as reward.

\begin{figure*}[htbp]
\includegraphics[width=0.9\linewidth]{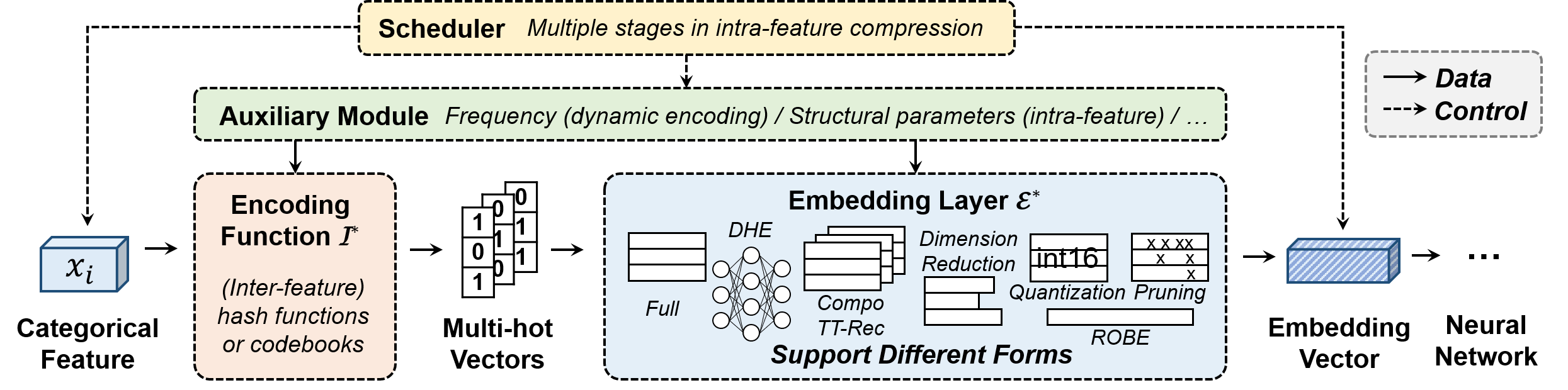}
\caption{Overview of the evaluation framework.}\label{fig:framework}
\end{figure*}

The above two methods adopt policy network to learn dimensions, incurring much training overhead for the trials of different settings. The memory budget can be considered in reward function just as NIS to enforce memory constraint at inference.

\noindent\textbf{AutoEmb}~\cite{DBLP:conf/icdm/ZhaoLFLTWCZLY21} forms MLP-based controllers, which take frequency and other contextual information as input and output probabilities of dimensions. Controllers and other model parameters are trained alternatively using DARTS~\cite{DBLP:journals/corr/abs-2204-00281} solution for bi-level optimization.

\noindent\textbf{AutoDim}~\cite{DBLP:conf/www/ZhaoLLTGSWGL21} defines field-wise architectural weights to compute probabilities of dimensions via gumbel-softmaxing. It also alternatively trains the architectural weights and the other model parameters using DARTS. After training, the dimension with the highest probability is selected for further re-training.

The above two methods both utilize DARTS, incurring lower training overhead than policy network. However, they cannot search within a given memory budget.

\noindent\textbf{AMTL} (Adaptively-Masked Twins-based Layer~\cite{DBLP:conf/cikm/YanWZLLXZ21}) introduces two MLPs for features with high- and low-frequency respectively, to output scores for positions where embeddings should be truncated.

\noindent\textbf{SSEDS} (Single-Shot Embedding Dimension Search~\cite{DBLP:conf/sigir/QuYTZSY22}) multiplies the pre-trained uncompressed embeddings with field-dimension-wise masks to conduct single-shot NAS. It uses the masks' gradients to represent value importance, which is further used to truncate the embeddings according to memory budget. 

\noindent\textbf{OptEmbed}~\cite{DBLP:conf/cikm/Lyu0ZG0TL22} jointly learns masks for both row and column. Row masks that threshold the embeddings' $L_1$ norms are multiplied onto the original embeddings for supernet training. After determining the row masks, OptEmbed conducts an evolutionary search to determine the column masks for embeddings truncation. 
Then the compressed embeddings are retrained to fit the masked parameters.

SSEDS and OptEmbed utilize one-shot NAS to make the training process faster.
While SSEDS takes the memory budget into consideration, OptEmbed does not support flexible memory budget.

In summary, dimension reduction methods aim to assign a suitable dimension for each feature.  
Except for MDE and AMTL, all methods require significant time for complex training or retraining. Except for AMTL and OptEmbed, other methods use projection matrices, which result in increased inference latency.
Only MDE, NIS and SSEDS can compress to a given memory budget. There are also other methods that jointly optimize the embedding dimension and model components using rule~\cite{DBLP:conf/iclr/ShenWGTWL21}, DARTS~\cite{DBLP:journals/corr/abs-2204-00281} or one-shot NAS~\cite{DBLP:conf/cikm/WeiW021}, which do not meet our plug-and-play requirement.

\subsubsection{Pruning}
Pruning is a common technique in the compression of deep learning models~\cite{DBLP:conf/nips/HanPTD15,DBLP:conf/iclr/0022KDSG17,DBLP:conf/iclr/FrankleC19}. According to the lottery ticket hypothesis~\cite{DBLP:conf/iclr/FrankleC19}, a dense neural network contains a subnetwork that can match the test accuracy of the dense network. Similar to dimension reduction that assigns different dimensions, pruning assigns different sparsity for different features. The pruned sparse embeddings are stored in sparse tensor format in practice. The symbol $r$ in Table~\ref{tab:intrafeature} means the density of embeddings.

\noindent\textbf{DeepLight}~\cite{DBLP:conf/wsdm/0002PZKFL21} uses structural pruning, a common pruning method in DL models~\cite{DBLP:journals/jetc/AnwarHS17}. It progressively thins out embeddings by filtering small-magnitude values, until reaching the memory budget. 

Except for DeepLight, all the others adopt learning methods, involving more training overhead for better model quality.

\noindent\textbf{PEP} (Plug-in Embedding Pruning~\cite{DBLP:conf/iclr/LiuGCJL21}) defines a learnable threshold for pruning.
After joint training the threshold and the other parameters, the model is retrained to fit the pruned embeddings. 

\noindent\textbf{HAM} (Hard Auxiliary Mask~\cite{DBLP:conf/cikm/XiaoXCC22}) first pre-trains the uncompressed embeddings with Soft Orthogonal~\cite{DBLP:conf/nips/BansalCW18} regularizations, then alternatively trains learnable masks and other parameters, and finally re-trains the pruned embeddings. 

\noindent\textbf{AutoSrh}~\cite{DBLP:journals/corr/abs-2006-04466,kong2022autosrh} sorts features by frequency and partitions them into blocks, with each block assigned with learnable masks for pruning. Masks and other model parameters are alternatively trained using DARTS. After training, parameters are filtered according to memory budget, then re-trained to fit the sparse embeddings.

The above methods learn masks or thresholds for pruning. HAM and AutoSrh update learnable masks alternatively with parameters, which is similar to SSEDS and OptEmbed in dimension reduction. Like dimension reduction, pruning attempts to allocate more memory to more important features.
Pruning methods achieve good model quality with significant training overhead. They are flexibly adapted to a given memory budget, but require system support for sparse tensor storage and computation.

\section{Evaluation Framework}\label{sec:framework}

We design and implement a unified modular evaluation framework for embedding compression, as shown in Figure~\ref{fig:framework}. 
Generally, all existing embedding compression methods can be implemented with these 4 modules: encoding function, embedding layer, scheduler, and auxiliary module. 
The encoding function inputs features and outputs one-hot or multi-hot vectors. 
The embedding layer stores embedding-related parameters, such as one or several embedding tables, MLPs, 1-D arrays, and sparse matrices, etc. For sparse matrices, we implement CSR and COO formats, and the framework adaptively chooses the format under a given memory budget. 
The embedding layer outputs the corresponding embeddings based on the encoded vectors, then
the embeddings are fed into neural networks along with numerical features for predictions. We omit the neural network part from the figure because it is not our focus. The optional auxiliary module contains data structures that assist model training, such as the frequency information in dynamic encoding, the learnable masks in pruning, the architecture weights in dimension reduction. The scheduler manages the entire training process, switches training stages, and schedules proper data to train certain parts of the model. For example, DARTS-based methods use training data and evaluation data to update model parameters and architecture weights respectively, while NAS-based methods usually require a re-training stage for further improvement.

The framework integrates 14 representative methods for experimental comparison, which are listed in Section~\ref{sec:expr:settings:method}. 
Besides existing methods, our framework supports any new method that applies this compression pipeline. 
We expect more compression methods to be proposed based on our framework.

The framework is implemented on Hetu~\cite{DBLP:journals/chinaf/MiaoNZZC23}, an efficient deep learning system. Our framework consists of 10 thousand lines of code in Python for the modules.
We also implement some necessary C++/CUDA computing kernels. 
The framework does not explicitly consider distributed scenarios: 
data parallelism can be simply applied, while model parallelism that partitions embedding layers are not necessary because embedding layers have already been compressed. 
Some other orthogonal system optimizations such as data prefetch~\cite{DBLP:conf/sigmod/MiaoSZZNY022}, or DL compilation~\cite{DBLP:conf/osdi/ChenMJZYSCWHCGK18} are not applied, as they do not affect our analysis.

\section{Experiments and Analysis}\label{sec:exp}
In this section, we experimentally evaluate the embedding compression methods on DLRM
(Section \ref{sec:expr:overall}).
We design experiments to reveal the influence of neural network models (Section~\ref{sec:expr:models}) and embedding dimensions (Section~\ref{sec:expr:dims}). We also apply the embedding compression methods to retrieval-augmented LLM
(Section~\ref{sec:expr:rag}). 
We later discuss challenges and future directions (Section~\ref{sec:expr:future}). 

    \subsection{Experiment Settings of DLRM}

\subsubsection{Models and Datasets.}
We experiment on three popular models: DLRM\footnote{In Section~\ref{sec:exp} we use the term DLRM to refer to this particular model, rather than the general deep learning recommendation models in the previous sections.}~\cite{DBLP:journals/corr/abs-1906-00091}, WDL~\cite{DBLP:conf/recsys/Cheng0HSCAACCIA16}, and DCN~\cite{DBLP:conf/kdd/WangFFW17}.
We evaluate three click-through rate (CTR) datasets: Avazu~\cite{avazu}, Criteo~\cite{criteo}, and Company, where the former two are widely used in academia and have been employed in recommendation benchmarks~\cite{mlperf,DBLP:conf/cikm/ZhuLYZH21}, and the latter is collected from a recommendation scenario in Tencent containing ad features.
The statistics of the datasets are listed in Table~\ref{tab:datasets}.

\begin{table}[htbp]
    \centering
    \small
    \caption{Overview of the datasets.}
    \begin{tabular}{c c c c}
    \toprule[1pt]
    Datasets & \# Fields & \# Features & \# Samples \\
    \midrule[0.8pt]
    Avazu & 22 & 9,449,445 & 40,428,967 \\
    Criteo & 26 & 33,762,577 & 45,840,617 \\
    Company & 43 & 66,102,027 & 35,682,429 \\
    \bottomrule[1pt]
    \end{tabular}
    \label{tab:datasets}
\end{table}

Feature frequency follows a power law~\cite{DBLP:conf/sigir/ZhangC15,DBLP:journals/corr/abs-2003-07336,DBLP:journals/pvldb/MiaoZSNYTC21,DBLP:conf/sigmod/MiaoSZZNY022,DBLP:conf/asplos/SethiAAKTW22,DBLP:conf/icde/ZhangCYYYZWDXSL22}.
For example, in Avazu and Criteo, the top 10\% features with the highest frequency account for more than 95\% of the occurrences in samples, while for the long-tail part, more than 80\% of the features have less than 5 occurrences.
Compression methods are inspired to allocate different amount of memory to features.

\subsubsection{Compared Methods}\label{sec:expr:settings:method}

We choose 14 representative methods for comparison. 
For static encoding, \textbf{CompoEmb} uses multiple hash functions, and DoubleHash, MEmCom can be regarded as its variants; \textbf{TT-Rec} is a special variant of CompoEmb that borrows the idea of tensor-train decomposition; \textbf{DHE} and \textbf{ROBE} explore different forms of embeddings; \textbf{Dedup} is the state-of-the-art similarity-based deduplication method.
For dynamic encoding, \textbf{MGQE} learns the codes of sub-embeddings and has a simpler PQ structure than LightRec; \textbf{AdaptEmb} employs feature frequency and is more general than another frequency-based method CEL.
For quantization, we choose \textbf{INT8/16} for fixed quantization with uniform value distribution; \textbf{ALPT} is the state-of-the-art method that learns the quantization scale.
For dimension reduction, \textbf{MDE} is the only heuristic-based method; \textbf{AutoDim} and \textbf{OptEmbed} are the state-of-the-art methods using trainable structural parameters and one-shot NAS respectively; we do not evaluate policy-gradient-based methods because they are time consuming and perform poorly.
For pruning, \textbf{DeepLight} is the only structural pruning method; \textbf{AutoSrh} is the state-of-the-art method that learns the pruning structure.

\subsubsection{Environment and Hyperparameters}

We use the Adam optimizer~\cite{DBLP:journals/corr/KingmaB14} with the learning rate grid-searched from [0.001, 0.01, 0.1], and the batch size is 64 (for Company) or 128 (for Avazu and Criteo). We conduct every single experiment on an Nvidia RTX TITAN 24 GB GPU card. 
We tested different dimensions on uncompressed embeddings, and selected 16 as the embedding dimension. 

For simplicity, we implement all methods on GPU. The location of the embeddings does not affect model accuracy or memory usage.
If the embedding layer is on CPU, embeddings need to be transferred to GPU with additional communications, and compute-intensive methods like TT-Rec and DHE which already have the highest latency will be slower. 
These two factors only affect the absolute value of processing time, not the relative ranking of each method.

\subsubsection{Metrics}
We employ AUC (area under the ROC curve) to measure model quality. 
In recommendation systems, an improvement of 0.001 in AUC is considerable. 
We measure memory usage by the actual memory consumption of the embedding layer at inference. This is more effective than using the number of parameters or sparsity rate, which do not account for the compression effect of quantization methods or the additional memory cost of sparse formats.
For training memory, we include the memory of the auxiliary modules. 
For training time, we measure the total time of training to convergence, including all stages. 
Inference latency is the forward pass time of a batch using well-trained model checkpoints.

\begin{table*}[ht]

\setlength{\arrayrulewidth}{1pt}
\newcommand{\rank}[1]{\scriptsize{\uline{#1}}}
\newcommand{\allline}[0]{\hhline{|-|-|-|-|-|-|-|-|-|-|-|-|-|}}
\newcommand{\leaveone}[0]{\hhline{|~|-|-|-|-|-|-|-|-|-|-|-|-|}}
\newcommand{\leavetwo}[0]{}
\newcommand{\catwidth}[0]{0.2cm}
\newcommand{\methodwidth}[0]{1.2cm}
\newcommand{\metricwidth}[0]{1.0cm}
\newcommand{\cwidth}[0]{0.93cm}
\newcommand{\companywidth}[0]{1.4cm}
\renewcommand{\arraystretch}{1.14}
\footnotesize
\caption{Overall performance on DLRM.}\label{table:overall}
\begin{tabular}{|p{\catwidth}<{\centering}|p{\methodwidth}<{\centering}|p{\metricwidth}<{\centering}|p{\cwidth}<{\centering}|p{\cwidth}<{\centering}|p{\cwidth}<{\centering}|p{\cwidth}<{\centering}|p{\cwidth}<{\centering}|p{\cwidth}<{\centering}|p{\cwidth}<{\centering}|p{\cwidth}<{\centering}|p{\companywidth}<{\centering}|p{\companywidth}<{\centering}|}
\allline
\multicolumn{2}{|c|}{} &  & \multicolumn{4}{c|}{Avazu} & \multicolumn{4}{c|}{Criteo} & \multicolumn{2}{c|}{Company} \\
\hhline{|~~|~|-|-|-|-|-|-|-|-|-|-|}
\multicolumn{2}{|c|}{\multirow{-2}{*}{Methods}} & \multirow{-2}{*}{Metrics} & 50\% & 10\% & 1\% & 0.1\% & 50\% & 10\% & 1\% & 0.1\% & 50\% & 1\% \\
\allline
\rowcolor{red!7} \cellcolor{white} & \cellcolor{white} &  AUC & \multicolumn{4}{c|}{ 0.7543 (100.0\%)} & \multicolumn{4}{c|}{ 0.8061 (100.0\%)} & \multicolumn{2}{c|}{0.7583 (100.0\%)} \\
\leavetwo
\rowcolor{blue!7} \cellcolor{white} & \cellcolor{white} & TrainMem &  \multicolumn{4}{c|}{ 300.0\%} & \multicolumn{4}{c|}{300.0\%} & \multicolumn{2}{c|}{300.0\%} \\
\leavetwo
\rowcolor{yellow!7} \cellcolor{white} & \cellcolor{white} & TrainTime &  \multicolumn{4}{c|}{ 1m36s}  & \multicolumn{4}{c|}{ 8m39s} & \multicolumn{2}{c|}{9m45s} \\
\leavetwo
\rowcolor{green!7} \multirow{-4}{*}{\cellcolor{white} \rotatebox[origin=c]{90}{Baseline}} & \multirow{-4}{*}{\cellcolor{white} Full} & Latency &  \multicolumn{4}{c|}{0.56ms}  & \multicolumn{4}{c|}{0.88ms} & \multicolumn{2}{c|}{0.61ms} \\
\allline
\rowcolor{red!7} \cellcolor{white} & \cellcolor{white} & AUC & 0.7491 & 0.7480 & 0.7472 & / & 0.8060 & \textbf{0.8053}\rank{3} & 0.8016 & / & 0.7503 & 0.7228 \\
\leavetwo
\rowcolor{blue!7} \cellcolor{white} & \cellcolor{white} & TrainMem & \textbf{150.0\%} & \textbf{30.0\%} & \textbf{3.0\%} & / & \textbf{150.0\%} & \textbf{30.0\%} & \textbf{3.0\%} & / & \textbf{150.0\%} & \textbf{3.0\%} \\
\leavetwo
\rowcolor{yellow!7} \cellcolor{white} & \cellcolor{white} & TrainTime & \textbf{6m42s}\rank{3} & \textbf{8m51s}\rank{2} & \textbf{5m23s}\rank{1} & / & 54m20s & \textbf{51m41s}\rank{3} & \textbf{50m27s}\rank{3} & / & 2h20m & \textbf{1h58m}\rank{3} \\
\leavetwo
\rowcolor{green!7} \cellcolor{white} & \multirow{-4}{*}{\cellcolor{white}CompoEmb} & Latency & 3.27ms & 3.34ms & 2.83ms & / & 4.47ms & 3.55ms & 3.92ms & / & 6.67ms & 6.57ms \\
\leaveone
\rowcolor{red!7} \cellcolor{white} & \cellcolor{white} & AUC & 0.7497 & 0.7537 & 0.7542 & 0.7517 & 0.7876 & 0.7993 & 0.8025 & 0.7997 & 0.7255 & 0.7329 \\
\leavetwo
\rowcolor{blue!7} \cellcolor{white} & \cellcolor{white} & TrainMem & \textbf{149.3\%} & \textbf{29.8\%} & \textbf{2.9\%} & \textbf{0.3\%} & \textbf{149.4\%} & \textbf{30.0\%} & \textbf{3.0\%} & \textbf{0.3\%} & \textbf{149.9\%} & \textbf{2.9\%} \\
\leavetwo
\rowcolor{yellow!7} \cellcolor{white} & \cellcolor{white} & TrainTime & 7h29m & 1h37m & 36m32s & \textbf{40m48s}\rank{3} & 12h8m & 12h4m & 3h29m & 4h30m & 127h27m & 5h53m \\
\leavetwo
\rowcolor{green!7} \cellcolor{white} & \multirow{-4}{*}{\cellcolor{white}TT-Rec} & Latency & 84.41ms & 21.28ms & 5.00ms & 4.28ms & 246.39ms & 82.72ms & 11.99ms & 6.10ms & 505.89ms & 24.05ms \\
\leaveone
\rowcolor{red!7} \cellcolor{white} & \cellcolor{white} & AUC & \textbf{0.7583}\rank{1} & \textbf{0.7581}\rank{1} & \textbf{0.7586}\rank{1} & \textbf{0.7563}\rank{1} & 0.8056 & \textbf{0.8053}\rank{3} & 0.8027 & 0.8004 & 0.7532 & \textbf{0.7442}\rank{2} \\
\leavetwo
\rowcolor{blue!7} \cellcolor{white} & \cellcolor{white} & TrainMem & \textbf{149.9\%} & \textbf{30.0\%} & \textbf{3.0\%} & \textbf{0.3\%} & \textbf{150.0\%} & \textbf{30.0\%} & \textbf{3.0\%} & \textbf{0.3\%} & \textbf{150.0\%} & \textbf{3.0\%} \\
\leavetwo
\rowcolor{yellow!7} \cellcolor{white} & \cellcolor{white} & TrainTime & 3h11m & 45m15s & 1h12m & 12h53m & 5h44m & 2h10m & 4h35m & 12h14m & 7h45m & 34h29m \\
\leavetwo
\rowcolor{green!7} \cellcolor{white} & \multirow{-4}{*}{\cellcolor{white}DHE} & Latency & 5.81ms & 3.37ms & 3.02ms & 5.91ms & 12.86ms & 6.52ms & 6.26ms & 8.58ms & 17.05ms & 20.35ms \\
\leaveone
\rowcolor{red!7} \cellcolor{white} & \cellcolor{white} & AUC & 0.6612 & 0.6522 & 0.6477 & 0.6407 & 0.7510 & 0.7479 & 0.7514 & 0.7490 & 0.5945 & 0.5828 \\
\leavetwo
\rowcolor{blue!7} \cellcolor{white} & \cellcolor{white} & TrainMem & \textbf{150.0\%} & \textbf{30.0\%} & \textbf{3.0\%} & \textbf{0.3\%} & \textbf{150.0\%} & \textbf{30.0\%} & \textbf{3.0\%} & \textbf{0.3\%} & \textbf{150.0\%} & \textbf{3.0\%} \\
\leavetwo
\rowcolor{yellow!7} \cellcolor{white} & \cellcolor{white} & TrainTime & 11m11s & 11m51s & 1h6m & 47m18s & 5h13m & 5h20m & 5h13m & 5h21m & \textbf{8m6s}\rank{1} & \textbf{8m4s}\rank{1} \\
\leavetwo
\rowcolor{green!7} \cellcolor{white} & \multirow{-4}{*}{\cellcolor{white}ROBE} & Latency & 0.66ms & \textbf{0.66ms}\rank{3} & \textbf{0.67ms}\rank{3} & 0.65ms & \textbf{0.92ms}\rank{1} & \textbf{0.87ms}\rank{1} & \textbf{0.95ms}\rank{2} & \textbf{0.91ms}\rank{2} & 0.67ms & 0.67ms \\
\leaveone
\rowcolor{red!7} \cellcolor{white} & \cellcolor{white} & AUC & \textbf{0.7547}\rank{2} & \textbf{0.7555}\rank{2} & \textbf{0.7573}\rank{3} & \textbf{0.7560}\rank{2} & \textbf{0.8076}\rank{1} & \textbf{0.8071}\rank{1} & \textbf{0.8057}\rank{3} & \textbf{0.8035}\rank{1} & \textbf{0.7553}\rank{3} & \textbf{0.7420}\rank{3} \\
\leavetwo
\rowcolor{blue!7} \cellcolor{white} & \cellcolor{white} & TrainMem & 300\% & 300\% & 300\% & 300\% & 300\% & 300\% & 300\% & 300\% & 300\% & 300\% \\
\leavetwo
\rowcolor{yellow!7} \cellcolor{white} & \cellcolor{white} & TrainTime & \textbf{2m44s}\rank{2} & \textbf{5m58s}\rank{1} & \textbf{9m54s}\rank{2} & \textbf{11m12s}\rank{1} & \textbf{10m10s}\rank{2} & \textbf{10m7s}\rank{1} & \textbf{18m48s}\rank{1} & \textbf{29m9s}\rank{1} & \textbf{10m1s}\rank{2} & \textbf{10m24s}\rank{2} \\
\leavetwo
\rowcolor{green!7} \multirow{-20}{*}{\cellcolor{white}\rotatebox[origin=c]{90}{Static Encoding}} & \multirow{-4}{*}{\cellcolor{white}Dedup} & Latency & 0.71ms & 0.82ms & 0.68ms & \textbf{0.64ms}\rank{2} & 1.29ms & \textbf{1.04ms}\rank{3} & \textbf{0.97ms}\rank{3} & \textbf{0.86ms}\rank{1} & 0.69ms & \textbf{0.65ms}\rank{3} \\
\allline
\rowcolor{red!7} \cellcolor{white} & \cellcolor{white} & AUC & \multicolumn{2}{c|}{0.7286 (16.0\%)} & \multicolumn{2}{c|}{0.7332 (11.1\%)} & \multicolumn{2}{c|}{0.7908 (16.1\%)} & \multicolumn{2}{c|}{0.7891 (11.2\%)} & 0.6881 (16.0\%) & 0.6477 (11.1\%) \\
\leavetwo
\rowcolor{blue!7} \cellcolor{white} & \cellcolor{white} & TrainMem & \multicolumn{2}{c|}{322.3\%} & \multicolumn{2}{c|}{317.4\%} & \multicolumn{2}{c|}{322.7\%} & \multicolumn{2}{c|}{317.8\%} & 322.2\% & 317.4\% \\
\leavetwo
\rowcolor{yellow!7} \cellcolor{white} & \cellcolor{white} & TrainTime & \multicolumn{2}{c|}{20m0s} & \multicolumn{2}{c|}{34m33s} & \multicolumn{2}{c|}{56m37s} & \multicolumn{2}{c|}{37m45s} & 3h36m & 1h13m \\
\leavetwo
\rowcolor{green!7} \cellcolor{white} & \multirow{-4}{*}{\cellcolor{white}MGQE} & Latency & \multicolumn{2}{c|}{3.40ms} & \multicolumn{2}{c|}{3.24ms} & \multicolumn{2}{c|}{4.33ms} & \multicolumn{2}{c|}{4.37ms} & 6.72ms & 6.74ms \\
\leaveone
\rowcolor{red!7} \cellcolor{white} & \cellcolor{white} & AUC & 0.7513 & 0.7484 & 0.7407 & 0.7302 & 0.8051 & 0.8026 & 0.7990 & 0.7921 & \textbf{0.7566}\rank{2} & 0.7136 \\
\leavetwo
\rowcolor{blue!7} \cellcolor{white} & \cellcolor{white} & TrainMem & 154.4\% & 35.3\% & 9.2\% & 6.5\% & 154.4\% & 35.3\% & 9.2\% & 6.5\% & 154.4\% & 9.2\% \\
\leavetwo
\rowcolor{yellow!7} \cellcolor{white} & \cellcolor{white} & TrainTime & 13m58s & \textbf{10m26s}\rank{3} & \textbf{13m13s}\rank{3} & 45m14s & 52m3s & \textbf{49m43s}\rank{2} & \textbf{48m36s}\rank{2} & \textbf{1h46m}\rank{2} & 2h7m & 2h11m \\
\leavetwo
\rowcolor{green!7} \multirow{-8}{*}{\rotatebox[origin=c]{90}{\cellcolor{white}Dynamic Encoding}} & \multirow{-4}{*}{\cellcolor{white}AdaptEmb} & Latency & 4.63ms & 3.59ms & 3.40ms & 3.13ms & 5.79ms & 4.40ms & 5.01ms & 4.55ms & 7.32ms & 7.30ms \\
\allline
\rowcolor{red!7} \cellcolor{white} & \cellcolor{white} & AUC & \textbf{0.7539}\rank{3} & \multicolumn{3}{c|}{0.7524 (25.0\%)} & \textbf{0.8071}\rank{2} & \multicolumn{3}{c|}{0.8045 (25.0\%)} & 0.7493 & 0.7536 (25.0\%) \\
\leavetwo
\rowcolor{blue!7} \cellcolor{white} & \cellcolor{white} & TrainMem & 250.0\% & \multicolumn{3}{c|}{225.0\%} & 250.0\% & \multicolumn{3}{c|}{225.0\%} & 250.0\% & 225.0\% \\
\leavetwo
\rowcolor{yellow!7} \cellcolor{white} & \cellcolor{white} & TrainTime & \textbf{2m24s}\rank{1} & \multicolumn{3}{c|}{4m31s} & \textbf{9m44s}\rank{1} & \multicolumn{3}{c|}{8m8s} & \textbf{10m16s}\rank{3} & 9m16s \\
\leavetwo
\rowcolor{green!7} \cellcolor{white} & \multirow{-4}{*}{\cellcolor{white}INT8/16} & Latency & \textbf{0.65ms}\rank{3} & \multicolumn{3}{c|}{0.65ms} & \textbf{1.01ms}\rank{3} & \multicolumn{3}{c|}{0.88ms} & \textbf{0.63ms}\rank{3} & 0.58ms \\
\leaveone
\rowcolor{red!7} \cellcolor{white} & \cellcolor{white} & AUC & \multicolumn{2}{c|}{0.7545 (56.3\%)} & \multicolumn{2}{c|}{0.7511 (31.3\%)} & \multicolumn{2}{c|}{0.8062 (56.3\%)} & \multicolumn{2}{c|}{0.8057 (31.3\%)} & 0.7562 (56.3\%) & 0.7532 (31.3\%) \\
\leavetwo
\rowcolor{blue!7} \cellcolor{white} & \cellcolor{white} & TrainMem & \multicolumn{2}{c|}{268.8\%} & \multicolumn{2}{c|}{243.8\%} & \multicolumn{2}{c|}{268.8\%} & \multicolumn{2}{c|}{243.8\%} & 268.8\% & 243.8\% \\
\leavetwo
\rowcolor{yellow!7} \cellcolor{white} & \cellcolor{white} & TrainTime & \multicolumn{2}{c|}{3m5s} & \multicolumn{2}{c|}{2m51s} & \multicolumn{2}{c|}{15m46s} & \multicolumn{2}{c|}{12m45s} & 16m28s & 17m22s \\
\leavetwo
\rowcolor{green!7} \multirow{-8}{*}{\cellcolor{white}\rotatebox[origin=c]{90}{Quantization}} & \multirow{-4}{*}{\cellcolor{white}ALPT} & Latency & \multicolumn{2}{c|}{0.63ms} & \multicolumn{2}{c|}{0.64ms} & \multicolumn{2}{c|}{0.92ms} & \multicolumn{2}{c|}{1.01ms} & 0.59ms & 0.61ms \\
\allline
\rowcolor{red!7} \cellcolor{white} & \cellcolor{white} & AUC & 0.7502 & \multicolumn{3}{c|}{0.7504 (8.1\%)} & 0.8044 & \multicolumn{3}{c|}{0.8033 (9.3\%)} & \textbf{0.7632}\rank{1} & / \\
\leavetwo
\rowcolor{blue!7} \cellcolor{white} & \cellcolor{white} & TrainMem & \textbf{150.0\%} & \multicolumn{3}{c|}{24.3\%} & \textbf{150.0\%} & \multicolumn{3}{c|}{27.9\%} & \textbf{150.0\%} & / \\
\leavetwo
\rowcolor{yellow!7} \cellcolor{white} & \cellcolor{white} & TrainTime & 7m28s & \multicolumn{3}{c|}{4m54s} & \textbf{33m48s}\rank{3} & \multicolumn{3}{c|}{34m58s} & 1h35m & / \\
\leavetwo
\rowcolor{green!7} \cellcolor{white} & \multirow{-4}{*}{\cellcolor{white}MDE} & Latency & 2.66ms & \multicolumn{3}{c|}{2.82ms} & 3.86ms & \multicolumn{3}{c|}{3.02ms} & 5.18ms & / \\
\leaveone
\rowcolor{red!7} \cellcolor{white} & \cellcolor{white} & AUC & \multicolumn{4}{c|}{0.7504 (46.4\%)} & \multicolumn{4}{c|}{0.8036 (46.0\%)} & \multicolumn{2}{c|}{0.7595 (37.3\%)} \\
\leavetwo
\rowcolor{blue!7} \cellcolor{white} & \cellcolor{white} & TrainMem & \multicolumn{4}{c|}{562.5\%} & \multicolumn{4}{c|}{562.5\%} & \multicolumn{2}{c|}{562.5\%} \\
\leavetwo
\rowcolor{yellow!7} \cellcolor{white} & \cellcolor{white} & TrainTime & \multicolumn{4}{c|}{1h6m} & \multicolumn{4}{c|}{5h13m} & \multicolumn{2}{c|}{2h27m} \\
\leavetwo
\rowcolor{green!7} \cellcolor{white} & \multirow{-4}{*}{\cellcolor{white}AutoDim} & Latency & \multicolumn{4}{c|}{1.11ms} & \multicolumn{4}{c|}{2.68ms} & \multicolumn{2}{c|}{5.27ms} \\
\leaveone
\rowcolor{red!7} \cellcolor{white} & \cellcolor{white} & AUC & \multicolumn{4}{c|}{0.7484 (50.0\%)} & \multicolumn{4}{c|}{0.8019 (46.3\%)} & \multicolumn{2}{c|}{0.7610 (47.0\%)} \\
\leavetwo
\rowcolor{blue!7} \cellcolor{white} & \cellcolor{white} & TrainMem & \multicolumn{4}{c|}{300.0\%} & \multicolumn{4}{c|}{300.0\%} & \multicolumn{2}{c|}{300.0\%} \\
\leavetwo
\rowcolor{yellow!7} \cellcolor{white} & \cellcolor{white} & TrainTime & \multicolumn{4}{c|}{16m17s} & \multicolumn{4}{c|}{30m34s} & \multicolumn{2}{c|}{1h57m} \\
\leavetwo
\rowcolor{green!7} \multirow{-12}{*}{\cellcolor{white}\rotatebox[origin=c]{90}{Dimension Reduction}} & \multirow{-4}{*}{\cellcolor{white}OptEmbed} & Latency & \multicolumn{4}{c|}{0.57ms} & \multicolumn{4}{c|}{0.79ms} & \multicolumn{2}{c|}{0.65ms} \\
\allline
\rowcolor{red!7} \cellcolor{white} & \cellcolor{white} & AUC & 0.7520 & \textbf{0.7554}\rank{3} & \textbf{0.7579}\rank{2} & \textbf{0.7526}\rank{3} & 0.8030 & 0.8050 & \textbf{0.8067}\rank{2} & \textbf{0.8019}\rank{3} & 0.7536 & \textbf{0.7504}\rank{1} \\
\leavetwo
\rowcolor{blue!7} \cellcolor{white} & \cellcolor{white} & TrainMem & 306.3\% & 306.3\% & 306.3\% & 306.3\% & 306.3\% & 306.3\% & 306.3\% & 306.3\% & 306.3\% & 306.3\% \\
\leavetwo
\rowcolor{yellow!7} \cellcolor{white} & \cellcolor{white} & TrainTime & 3h32m & 3h45m & 5h25m & 8h10m & 3h11m & 8h41m & 15h38m & 15h41m & 31h25m & 27h50m \\
\leavetwo
\rowcolor{green!7} \cellcolor{white} & \multirow{-4}{*}{\cellcolor{white}DeepLight} & Latency & \textbf{0.59ms}\rank{2} & \textbf{0.63ms}\rank{2} & \textbf{0.62ms}\rank{2} & \textbf{0.64ms}\rank{2} & \textbf{0.93ms}\rank{2} & \textbf{0.91ms}\rank{2} & \textbf{0.91ms}\rank{1} & \textbf{0.93ms}\rank{3} & \textbf{0.56ms}\rank{1} & \textbf{0.58ms}\rank{1} \\
\leaveone
\rowcolor{red!7} \cellcolor{white} & \cellcolor{white} & AUC & 0.7518 & 0.7520 & 0.7526 & 0.7507 & \textbf{0.8066}\rank{3} & \textbf{0.8071}\rank{1} & \textbf{0.8068}\rank{1} & \textbf{0.8033}\rank{2} & 0.7529 & 0.7215 \\
\leavetwo
\rowcolor{blue!7} \cellcolor{white} & \cellcolor{white} & TrainMem & 306.3\% & 306.3\% & 306.3\% & 306.3\% & 306.3\% & 306.3\% & 306.3\% & 306.3\% & 306.3\% & 306.3\% \\
\leavetwo
\rowcolor{yellow!7} \cellcolor{white} & \cellcolor{white} & TrainTime & 22m36s & 28m43s & 29m55s & \textbf{32m39s}\rank{2} & 2h9m & 2h2m & 2h6m & \textbf{2h9m}\rank{3} & 4h21m & 4h43m \\
\leavetwo
\rowcolor{green!7} \multirow{-8}{*}{\cellcolor{white}\rotatebox[origin=c]{90}{Pruning}} & \multirow{-4}{*}{\cellcolor{white}AutoSrh} & Latency & \textbf{0.56ms}\rank{1} & \textbf{0.57ms}\rank{1} & \textbf{0.57ms}\rank{1} & \textbf{0.62ms}\rank{1} & 1.43ms & 1.36ms & 1.49ms & 1.44ms & \textbf{0.57ms}\rank{2} & \textbf{0.59ms}\rank{2} \\
\allline
\end {tabular}
\end {table*}

    \subsection{Performance on DLRM}\label{sec:expr:overall}

Table~\ref{table:overall} shows the results on DLRM under different inference memory budgets. 
We use the uncompressed embedding table as the baseline method, and its memory usage as the baseline memory usage.
By default, the inference memory budgets are 50\%, 10\%, 1\%, 0.1\% of the baseline memory.
Methods that cannot achieve these compression ratios are compressed as much as possible, with their actual memory usage listed in parentheses; these results are not explicitly compared with those normal ones.

\subsubsection{Ability of Compression.}

\textbf{Hash-based methods (including LSH) and pruning methods are the most capable compression methods, achieving all compression ratios.} 
Static encoding methods and AdaptEmb can simply adjust the number of rows, while pruning methods can flexibly change the sparsity. 
The storage of auxiliary mapping in LSH-based Dedup can be reduced by using large-sized tensor blocks.
All the other methods have certain limitations on their compression capabilities.
Other dynamic encoding methods need to store feature-to-embedding mappings with memory proportional to the number of features, leading to an upper bound of $16\times$ compression ratio. Quantization methods are limited to several specific compression ratios: $2\times$ and $4\times$ for simple INT8/16; $1.8\times$ and $3.2\times$ for ALPT which requires more memory for feature-wise step sizes. Dimension reduction methods learn optimal dimensions based on model quality rather than memory budgets: MDE and AutoDim assign features with at least one dimension, leading to an upper bound of $16\times$ compression ratio; AutoDim and OptEmbed achieve specific compression ratios within $2-2.7\times$.

\subsubsection{Model AUC}

\textbf{In general, static encoding, quantization, and pruning methods achieve the best model AUC.}
For each memory budget, the methods with top-3 AUC are highlighted in bold, coupled with an underlined ranking. There is no single method that performs best in all situations.
Specifically, quantization methods perform well around 25\% or 50\% of the baseline memory, since they do not change the original training paradigm.
DHE adopts novel MLP structures and performs well on Avazu dataset.
Pruning methods use more memory to emphasize important features, regardless of memory budgets, and they perform well on Criteo dataset, especially when the memory budget is small.
Dedup achieves near-optimal performance on all datasets, demonstrating the strength of similarity-based deduplication.
Dimension reduction methods can also achieve good model AUC by capturing feature importance, but the results are mostly not comparable due to different compression ratios.
Which methods are suitable for different datasets remains an open question, which we leave as future work.

Not all methods achieve better AUC with larger memory budgets. 
For TT-Rec and DHE, the dimension of matrix multiplication increases as the memory increases, making the optimization more difficult. 
For Dedup, within small memory, pooly-trained embeddings may be replaced by well-trained ones, thus improving model quality. 
For MGQE, larger memory only means the embeddings are split into more parts, with the centroids memory unchanged. 
For pruning, noisy redundant parameters may be removed as the memory decreases, leading to an increase in AUC. 

\begin{figure*}[htbp]
    \centering
    \includegraphics[width=\linewidth]{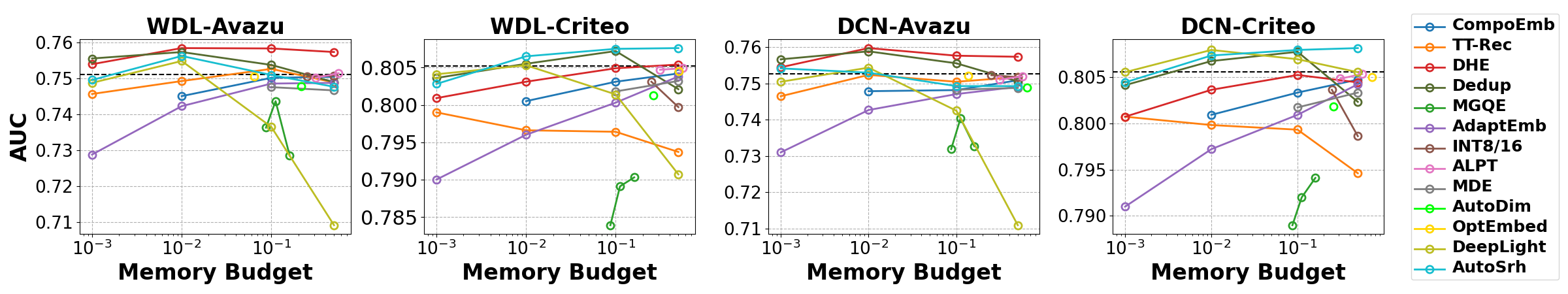}
    \caption{AUC of WDL and DCN.}
    \label{fig:aucs}
\end{figure*}

\subsubsection{Training Memory.}

\textbf{Static encoding methods (except for Dedup) and MDE use the least memory during training, only three times the inference memory budget considering the optimizer states.}
Memory consumption during training is different from inference. Many methods require training uncompressed embeddings or other memory-intensive auxiliary modules. 
For the Adam optimizer, we also need to store the first- and the second-order momentum, making the memory usage at least three times that of the inference process. 
Training memory is described as a ratio of the baseline memory, independent of dataset size.

Static encoding methods (except for Dedup) and MDE have no extra structures, and the trained parameters are directly used for inference. They have a linear relationship between training memory and inference memory, where the exact multiple depends on the optimizer. 
AdaptEmb records feature frequency during training.
Quantization only quantizes the embeddings, not the optimizer states, so its training memory is large. MGQE, Dedup, dimension reduction methods (except MDE) and pruning methods, require full-embedding training which is at least three times the baseline memory. Among them, AutoDim requires the most memory because it simultaneously trains all candidate dimensions.

\subsubsection{Training Time.} 

\textbf{Simple hash-based methods (including Dedup) and INT8/16 are fast to converge.}
We employ the early stopping strategy and record the time for each method to converge, including all stages. Methods with top-3 least training time are highlighted in bold with an underlined ranking. Generally speaking, the larger the dataset, the longer it takes for DLRM to converge.

Dedup and INT8/16 are the fastest to converge. They do not change the training paradigm and involve negligible deduplication and (de)quantization overhead.
CompoEmb and AdaptEmb are also fast to converge, requiring minor modifications to the training process. Other inter-feature compression methods either involve complex computations, or require more epochs to converge due to relaxed abstraction of embedding tables.
These also result in large variance in their training times.
Dimension reduction and pruning methods have longer training times due to the introduction of warm-up, search, retraining stages, and the alternative learning of model parameters and structural parameters.

There is not a clear relationship between the training time and the memory budget. On the one hand, more memory may lead to greater training complexity; on the other hand, less memory may make it harder to achieve convergence. 

\subsubsection{Inference Latency.}

\textbf{Dedup, ROBE, OptEmbed, quantization methods, and pruning methods have the lowest inference latency.}
After training, the model checkpoints are saved for inference. 
The methods with top-3 least inference latency are highlighted in bold with an underlined ranking. We use the same batch size in training and inference.
Criteo has greater latency than Avazu with more embeddings to compute. The batch size of Company is smaller than other datasets, so the latency is not comparable.

Dedup, ROBE, OptEmbed, quantization, and pruning all lookup the embeddings from only one table (or array), resulting in low inference latency. Dedup conducts similarity-based deduplication, with no need to consider field information; ROBE designs an array to share all embeddings. Except for Dedup and ROBE, inter-feature compression methods have to perform compression within fields, considering that features of the same field have similar semantics. 
TT-Rec and DHE have the largest inference latency due to time-consuming matrix multiplications. Quantization only incurs negligible dequantization process during inference. Sparse tensors in pruning may have fewer memory accesses with no additional overheads. MDE and AutoDim introduce additional matrix multiplications to align dimensions, thereby increasing inference latency. 

If the time complexity is constant, the inference latency hardly changes with the memory budget.
In contrast, TT-Rec and DHE perform more complex computations with larger memory, resulting in greater latency. However, when the memory is too small, more fields participate in compression, also leading to greater latency.

\subsubsection{Commercial Dataset}
\textbf{After analyzing the results on the commercial dataset Company, we find that the conclusions are consistent with those of the public datasets, despite some minor differences.} The training time variance is larger on Company, mainly because the larger Company dataset is more difficult for compression methods to train. Compression methods perform similarly on the three datasets, because 1) the public datasets are also collected from real recommendation scenarios; 2) our conclusions are robust enough to be generalized to larger datasets. Since we use compression ratios to study the performance, the absolute size of the embedding table has little impact on the conclusions.

\subsubsection{Discussion on Taxonomy}
\textbf{The current taxonomy is based on the compression paradigm, which determines the implementation.} For example, dynamic encoding records dynamic mappings, quantization adopts low-precision data types, and pruning stores embeddings in sparse formats. In experiments, methods of the same category have a certain degree of similarity, but there may be differences in some metrics due to different techniques used, such as simple hashing, complex computation, similarity-based deduplication, and VQ techniques in inter-feature compression, heuristics, policy gradient, DARTS, and one-shot-NAS techniques in intra-feature compression.
Our analysis considers both paradigms and techniques, making the conclusions more comprehensive.

    \subsection{Impact of Neural Network Model}\label{sec:expr:models}

For another two recommendation models WDL~\cite{DBLP:conf/recsys/Cheng0HSCAACCIA16} and DCN~\cite{DBLP:conf/kdd/WangFFW17}, we plot the AUC of each compression method at each inference memory budget in Figure~\ref{fig:aucs}. Despite minor differences compared to DLRM,
the ranking of the methods remains almost the same. 
The neural networks’ memory consumption and processing time have minor differences and do not impact the conclusions.

From the experimental results, we can see that the optimization of the model and the selection of the compression method are orthogonal, as the compression methods are decoupled from the downstream neural network. Therefore, the conclusions we draw on DLRM can be applied to other models as well.

\begin{figure}[htbp]
\centering
\begin{minipage}[t]{0.6\linewidth}
\centering
\includegraphics[width=\linewidth]{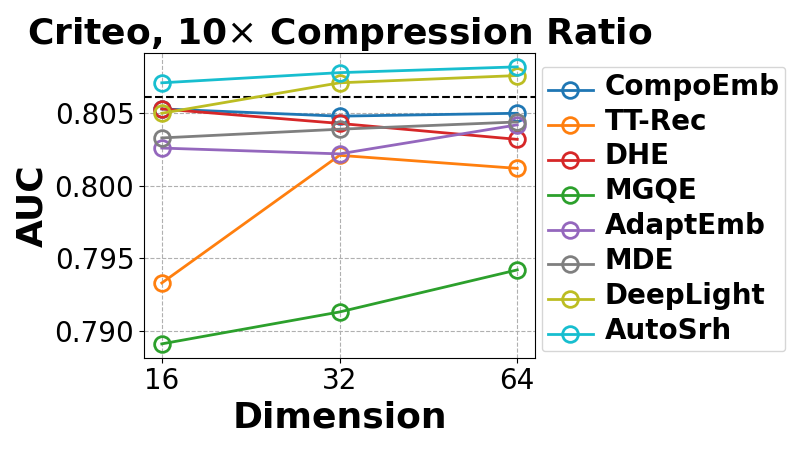}
\caption{AUC vs dimension.}\label{fig:dimension}
\end{minipage}%
\begin{minipage}[t]{0.375\linewidth}
\centering
\includegraphics[width=\linewidth]{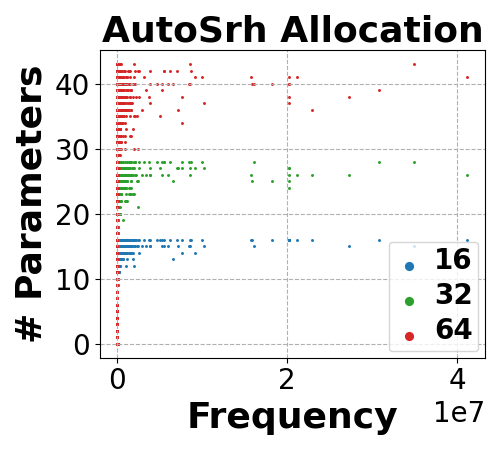}
\caption{Allocation.}\label{fig:density}
\end{minipage}%
\centering
\end{figure}

    \subsection{Impact of Dimension}\label{sec:expr:dims}

In Section~\ref{sec:expr:overall}, we align the embedding dimension to the baseline. However, methods that capture feature frequency or importance generally prefer larger dimensions to allocate more memory for more important features ~\cite{DBLP:journals/corr/abs-2212-05735,DBLP:conf/kdd/ChenYZHWW21,DBLP:conf/cikm/Lyu0ZG0TL22,DBLP:conf/sigir/QuYTZSY22,DBLP:conf/iclr/LiuGCJL21,kong2022autosrh}. In this section, we enlarge the dimension to explore the potential of these methods. 

Figure~\ref{fig:dimension} shows the AUC for each compression method with dimension 16, 32, and 64. The inference memory budget is fixed at 10\% of the baseline memory with dimension 16.
In general, methods that adopt feature importance, including dynamic encoding, dimension reduction, and pruning, have a certain increase in AUC as the dimension increases. In contrast, static encoding methods mostly do not benefit from larger dimensions.

In Figure~\ref{fig:density}, we visualize the actual memory allocated for each feature in AutoSrh. 
Each point represents a feature: the x-axis is its frequency, and the y-axis is the number of assigned parameters.
The allocated memory does not necessarily depend on frequency, as frequency is only one factor of feature importance.
As the dimension becomes larger, AutoSrh allocates more memory to important features, explaining the effect of dimension increase.

When the dimension is enlarged, MGQE and pruning methods increase the training memory linearly despite better AUC. Therefore, choosing the right dimensions requires a careful trade-off between model quality and training overhead.


    \subsection{Performance on Retrieval-augmented LLM}\label{sec:expr:rag}

In this section, we apply compression methods to generated embeddings in a retrieval-augmented LLM. 
The entire generated embeddings are produced by neural networks and present only at inference, different from parametric embeddings in DLRM that are trainable parameters and present throughout training.
Therefore, compression methods that involve the training process, such as AutoML-based methods, are not suitable for generated embeddings.

We select applicable compression methods or their variants for evaluation, including \textbf{TT} (tensor-train decomposition), \textbf{Dedup}, \textbf{PQ}, \textbf{MagPQ} (PQ within embedding groups that are split by magnitude), \textbf{INT8/16}, \textbf{SVD} (dimension reduction), \textbf{MagSVD} (SVD within embedding groups that are split by magnitude), \textbf{Pruning} (pruning values of low magnitude). MagPQ and MagSVD are variants of MGQE and MDE respectively, replacing missing frequency information with embeddings' L2-norms.

We experiment with RAG~\cite{DBLP:conf/nips/LewisPPPKGKLYR020} which uses DPR~\cite{DBLP:conf/emnlp/KarpukhinOMLWEC20} for retrieval and BART~\cite{DBLP:conf/acl/LewisLGGMLSZ20} for generation.
We experiment on the open-domain QA dataset Natural Questions (NQ)~\cite{DBLP:journals/tacl/KwiatkowskiPRCP19}, with cleaned Wikipedia articles (21 million) as the search corpus, following previous research~\cite{DBLP:conf/emnlp/KarpukhinOMLWEC20,DBLP:conf/naacl/QuDLLRZDWW21,DBLP:conf/emnlp/GaoC21,DBLP:conf/nips/LewisPPPKGKLYR020,DBLP:journals/corr/abs-2309-13335}. 
We retrieve top 10 documents for each query. The embedding dimension is 768, which is much larger than DLRM.
Since the entire embeddings are generated after training, we apply compression methods at the inference stage. 
Each experiment is conducted on an Nvidia A100 40GB GPU card. 
Table~\ref{table:rag} presents three metrics: Exact Match (EM) score, compression time, and batched-decompression latency with a batch size of 1024.

\subsubsection{Ability of Compression}
\textbf{TT, Dedup, and Pruning can reach all compression ratios.} Similar to the DLRM experiment, we compress under four memory budgets. 
TT essentially performs two SVDs with moderate dimensions to enable a wide range of compression ratios.
In contrast, (Mag)SVD cannot support small memory budgets, because their memory scales linearly with the corpus cardinality. Pure SVD cannot support large memory budgets, because it is difficult to decompose with large intermediate dimensions.
Dedup and Pruning have adjustable thresholds, making them applicable for almost any memory budget. (Mag)PQ cannot support large compression ratios due to complex computations in clustering. INT8/16 only support several fixed compression ratios.

\subsubsection{Embedding Quality}
\textbf{INT8/16, (Mag)SVD, Pruning achieve best EM scores under large memory budgets, while (Mag)PQ achieve best EM scores under small memory budgets.} INT8/16 and PQ have been implemented in the well-known embedding search library Faiss~\cite{DBLP:journals/tbd/JohnsonDJ21} due to their effectiveness. MagPQ and MagSVD show comparable performance to PQ and SVD with less memory, thanks to magnitude-aware compression. 
TT uses two SVDs, which greatly degrades performance. Dedup's block-wise deduplication may not perform well on this retrieval-related task.

\subsubsection{Compression Time}
\textbf{INT8/16 has the smallest compression time, followed by Pruning.} INT8/16 requires almost no computation. Pruning uses an efficient binary search algorithm to determine the threshold. Dedup uses L2LSH for deduplication which is only efficient under large memory budgets when the block size is large. (Mag)SVD is only efficient under small memory budgets when the intermediate dimensions are small. TT and (Mag)PQ are computationally expensive, resulting in long compression times.

\subsubsection{Batched-decompression Latency}
\textbf{INT8/16 has the smallest latency, followed by Pruning, Dedup, and (Mag)PQ.} 
The decompression of INT8/16 and CSR-format Pruning is fast with little overhead. 
When the memory budget is small, Pruning uses COO format, which is very slow for high-dimensional embeddings.
Dedup and (Mag)PQ look up embeddings from tensor blocks or centroids, with no computation overhead. TT and (Mag)SVD adopt matrix multiplication, leading to large decompression latency.

    \subsection{Further Discussions and Future Directions}\label{sec:expr:future}

\subsubsection{Challenges}

Currently, all compression methods have certain drawbacks, requiring users to carefully trade-off based on practical needs. 
For DLRM, there is no single method that performs well in all metrics.
For retrieval-augmented LLM, research on embedding compression is still in its early stages, with only a few specialized methods available.
Therefore, more comprehensive and advanced methods are expected in both fields.

On the other hand, the relationship between datasets and compression methods has not been studied. Our experiments show that different methods perform better on different datasets in DLRM, but it is unclear why. It is currently difficult to determine a proper method for a given dataset without actual experiments.

\begin{table}[ht]
\setlength{\arrayrulewidth}{1pt}
\newcommand{\rank}[1]{\scriptsize{\uline{#1}}}
\newcommand{\allline}[0]{\hhline{|-|-|-|-|-|-|}}
\newcommand{\leaveone}[0]{}
\newcommand{\cwidth}[0]{1.05cm}
\small
\caption{Overall performance on RAG.}\label{table:rag}
\begin{tabular}{|p{1cm}<{\centering}|p{0.82cm}<{\centering}|p{\cwidth}<{\centering}|p{\cwidth}<{\centering}|p{\cwidth}<{\centering}|p{\cwidth}<{\centering}|}
\allline
\textbf{Methods} & \textbf{Metrics} & \textbf{50\%} & \textbf{10\%} & \textbf{1\%} & \textbf{0.1\%} \\
\allline
\rowcolor{red!7} \cellcolor{white} Full &  EM & \multicolumn{4}{c|}{41.14 (100.00\%)} \\
\allline
\rowcolor{red!7} \cellcolor{white} & EM & 22.02 & 11.47 & 6.54 & 4.38 \\
\leaveone
\rowcolor{yellow!7} \cellcolor{white} & Time & 1h35m & 42m46s & 18m11s & 13m22s \\
\leaveone
\rowcolor{green!7} \multirow{-3}{*}{\cellcolor{white}TT} & Latency & 39.54s & 9.14s & 2.21s & 482.77ms \\
\allline
\rowcolor{red!7} \cellcolor{white} & EM & 25.93 & 15.57 & 5.51 & 4.68 \\
\leaveone
\rowcolor{yellow!7} \cellcolor{white} & Time & 11m3s & 11m29s & 33m36s & 2h34m \\
\leaveone
\rowcolor{green!7} \multirow{-3}{*}{\cellcolor{white}Dedup} & Latency & 4.14ms & 3.55ms & 3.00ms & 2.63ms \\
\allline
\rowcolor{red!7} \cellcolor{white} & EM & \multicolumn{2}{c|}{35.32 (3.14\%)} & \multicolumn{2}{c|}{24.57 (1.58\%)} \\
\leaveone
\rowcolor{yellow!7} \cellcolor{white} & Time & \multicolumn{2}{c|}{43m21s} & \multicolumn{2}{c|}{40m33s}  \\
\leaveone
\rowcolor{green!7} \multirow{-3}{*}{\cellcolor{white}PQ} & Latency & \multicolumn{2}{c|}{8.29ms} & \multicolumn{2}{c|}{9.09ms}  \\
\allline
\rowcolor{red!7} \cellcolor{white} & EM & \multicolumn{2}{c|}{35.29 (2.99\%)} & \multicolumn{2}{c|}{33.99 (1.57\%)}  \\
\leaveone
\rowcolor{yellow!7} \cellcolor{white} & Time & \multicolumn{2}{c|}{1h9m} & \multicolumn{2}{c|}{40m30s}  \\
\leaveone
\rowcolor{green!7} \multirow{-3}{*}{\cellcolor{white}MagPQ} & Latency & \multicolumn{2}{c|}{11.52ms} & \multicolumn{2}{c|}{10.71ms} \\
\allline
\rowcolor{red!7} \cellcolor{white} & EM & 41.02 & \multicolumn{3}{c|}{38.06 (25.00\%)}  \\
\leaveone
\rowcolor{yellow!7} \cellcolor{white} & Time & 4m60s & \multicolumn{3}{c|}{5m3s} \\
\leaveone
\rowcolor{green!7} \multirow{-3}{*}{\cellcolor{white}INT8/16} & Latency & 0.0513ms & \multicolumn{3}{c|}{0.0472ms}  \\
\allline
\rowcolor{red!7} \cellcolor{white} & EM & / & 31.02 & 3.88 & /  \\
\leaveone
\rowcolor{yellow!7} \cellcolor{white} & Time & / & 25m48s & 10m4s & / \\
\leaveone
\rowcolor{green!7} \multirow{-3}{*}{\cellcolor{white}SVD} & Latency & / & 16.86ms & 11.58ms & / \\
\allline
\rowcolor{red!7} \cellcolor{white} & EM & 41.11 & 30.89 & 4.04 & / \\
\leaveone
\rowcolor{yellow!7} \cellcolor{white} & Time & 50m6s & 17m16s & 13m0s & / \\
\leaveone
\rowcolor{green!7} \multirow{-3}{*}{\cellcolor{white}MagSVD} & Latency & 65.83ms & 35.56ms & 25.57ms & / \\
\allline
\rowcolor{red!7} \cellcolor{white} & EM & 37.29 & 17.15 & 4.24 & 4.04 \\
\leaveone
\rowcolor{yellow!7} \cellcolor{white} & Time & 14m37s & 11m2s & 12m13s & 13m25s \\
\leaveone
\rowcolor{green!7} \multirow{-3}{*}{\cellcolor{white}Pruning} & Latency & 3.13ms & 3.42ms & 3.23ms & 1.33s \\
\allline
\end {tabular}
\end {table}

\subsubsection{Future Directions}

A straightforward idea is to combine the advantages of different compression methods in DLRM.
For dynamic encoding, state-of-the-art static encoding and pruning methods can be integrated to achieve better model quality in online scenarios.
Quantization can be used as a plug-in module, contributing a fixed compression ratio with very low cost; another possible improvement is to assign data types with different bits to different features, borrowing ideas of capturing feature importance.
Dimension reduction and pruning require pre-training, where
static encoding can be applied to avoid large training memory.

Currently, embedding compression for retrieval tasks mainly uses quantization or PQ~\cite{DBLP:journals/tbd/JohnsonDJ21}. To the best of our knowledge, we are the first to study other embedding compression methods for retrieval. 
We anticipate that compression methods specifically designed for retrieval will emerge in the future and can be combined with embedding search to further improve performance.
Inspired by data skewness in DLRM, we are also curious whether retrieval datasets also have such properties, which we leave as future work.

Moreover, studying the impact of recommendation data distribution on compression methods is also a promising direction. 
At present, for a given dataset we can only determine compression methods experimentally.
A deeper understanding of data will not only help in the selection of compression methods, but also inspire the development of more advanced methods.

\section{Conclusion}\label{sec:conclude}
In this paper, we surveyed existing embedding compression methods and proposed a new taxonomy. We modularized the compression pipeline and implemented a unified evaluation framework. We conducted a comprehensive experimental evaluation to analyze the performance of each method under different memory budgets. The experimental results reveal the pros and cons of each method, provide suggestions for method selection in different situations, and shed light on promising research directions.

\begin{acks}
 This work is supported by National Key R\&D Program of China (2022ZD0116315), National Natural Science Foundation of China (U22B2037 and U23B2048), and PKU-Tencent joint research Lab.
 Yingxia Shao's work is supported by the National Natural Science Foundation of China (Nos. 62272054, 62192784), Beijing Nova Program (No. 20230484319), the Fundamental Research Funds for the Central Universities (No. 2023PY11) and Xiaomi Young Talents Program.
  Bin Cui and Xupeng Miao are the co-corresponding authors.
\end{acks}

\clearpage
\balance

\normalem
\bibliographystyle{ACM-Reference-Format}
\bibliography{reference}



\end{document}